\documentclass[journal]{IEEEtran}
\usepackage{ifpdf}

\usepackage{cite}

\ifCLASSINFOpdf
\else
\fi
\usepackage{amsmath,amssymb}
\usepackage{algorithmic}

\usepackage{array}

\usepackage{stfloats}
\usepackage{url}

\usepackage{multicol}
\usepackage[bookmarks=true]{hyperref}

\usepackage{algorithm2e}
\usepackage[english]{babel}
\usepackage{color}
\usepackage{epic}
\usepackage{epsfig}
\usepackage{epstopdf}
\usepackage[multiple]{footmisc}
\usepackage{graphicx}
\graphicspath{{figures/}}
\usepackage{placeins}
\usepackage{wrapfig}
\usepackage[dvipsnames]{xcolor}

\begin{document}

%
\title{Distributed Learning of Decentralized Control Policies for Articulated Mobile Robots}
%
%
%

\author{Guillaume~Sartoretti,~\IEEEmembership{Member,~IEEE,}
        William~Paivine,~\IEEEmembership{Student~Member,~IEEE,} \\
        Yunfei~Shi,~\IEEEmembership{Student~Member,~IEEE,}
        Yue~Wu,~\IEEEmembership{Student~Member,~IEEE,}
        and~Howie~Choset,~\IEEEmembership{Fellow,~IEEE}
\thanks{G. Sartoretti, H. Choset, W. Paivine, and Y. Wu are with the Robotics Institute at Carnegie Mellon University, Pittsburgh, PA 15213, USA. {\tt\small \{gsartore,wjp,ywu5,choset\}@andrew.cmu.edu}.}
\thanks{Y. Shi is with the Department of Electrical Engineering at The Hong Kong Polytechnic University, Hong Kong. {\tt\small yunfei.shi@connect.polyu.hk}.}
\thanks{Manuscript received October 28, 2018;}}

%
%

\markboth{IEEE Transactions in Robotics, June~2019}%
{Sartoretti \MakeLowercase{\textit{et al.}}: Distributed Learning of Decentralized Control Policies for Articulated Mobile Robots}
%



\maketitle

\begin{abstract}

State-of-the-art distributed algorithms for reinforcement learning rely on multiple independent agents, which simultaneously learn in parallel environments\footnote{Independent copies of the same learning environment, i.e., game simulation in which a reinforcement learning agent is allowed to explore its state-action space, collect rewards, and ultimately learn a policy.}
while asynchronously updating a common, shared policy.
Moreover, decentralized control architectures (e.g., CPGs) can coordinate spatially distributed portions of an articulated robot to achieve system-level objectives.
In this work,
we investigate the relationship between distributed learning and decentralized control by learning decentralized control policies for the locomotion of articulated robots in challenging environments.
To this end, we present an approach that leverages the structure of the asynchronous advantage actor-critic (A3C) algorithm to provide a natural means of learning decentralized control policies on a single articulated robot.
Our primary contribution shows individual agents in the A3C algorithm can be defined by independently controlled portions of the robot's body, thus enabling distributed learning on a single robot for efficient hardware implementation.
We present results of closed-loop locomotion in unstructured terrains on a snake and a hexapod robot, using decentralized controllers learned offline and online respectively, as a natural means to cover the different key applications of our approach.
For the snake robot, we are optimizing the forward progression in unstructured environments, but for the hexapod robot, the goal is to maintain a stabilized body pose.
Our results show that the proposed approach can be adapted to many different types of articulated robots by controlling some of their independent parts in a distributed manner, and the decentralized policy can be trained with high sample efficiency.

\end{abstract}


%
\IEEEpeerreviewmaketitle

\section{Introduction}

\IEEEPARstart{A}{rticulated} mobile robots such as snake and legged robots have the potential to excel at locomoting through a large variety of environments by leveraging their ability to adapt their shape to their surrounding terrain~\cite{Tesch2009}.
In the case of snake robots in particular, recent results have shown that a decentralized control architecture can improve locomotion through unstructured and cluttered environments~\cite{Whitman2016,Travers2015,kano2012local}.
For articulated robots, in general, decentralized control is generally achieved by partitioning the robots' body into spatially distributed portions, which can then be coordinated by a set of coupled central pattern generators (CPGs)~\cite{ijspeert2008central,crespi2006amphibot}.
In deep reinforcement learning (RL), new state-of-the-art methods are also based on the idea of distributing the learning task to several agents, which can then asynchronously update a common, global policy.
The policy is generally represented by a function, mapping the current state of an agent to the optimal action the agent should enact; this function is very often approximated by a neural network~\cite{mnih2013playing,van2016deep,mnih2016asynchronous}.
In this paper, we present a learning approach that leverages the general structure of asynchronous distributed learning algorithms~\cite{mnih2016asynchronous} as a natural means to learn decentralized control policies for a single articulated platform.
In particular, we focus on the A3C algorithm among all asynchronous distributed learning algorithms~\cite{mnih2016asynchronous} because it allows agents to learn a stochastic policy (rather than a deterministic one such as a Q-table).
Stochastic policies are usually more robust to the many sources of noise and uncertainty that are integral to any hardware training and deployment~\cite{haarnoja2017reinforcement}.

\begin{figure}[t]
\vspace{0.05cm}
\begin{center}
\includegraphics[width=0.97\linewidth]{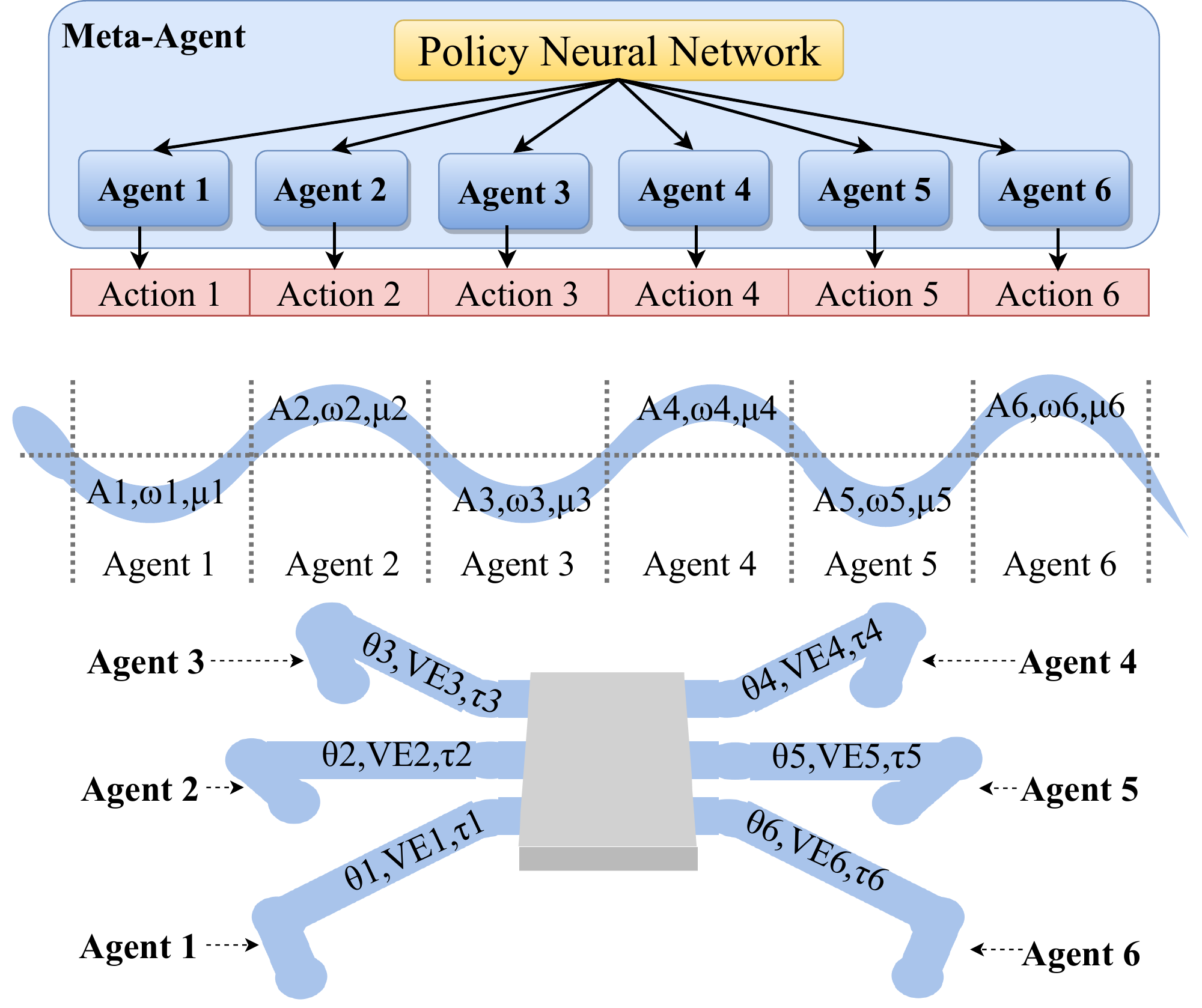}
\end{center}
\vspace{-0.65cm}
\caption{A3C meta-agent containing multiple learning agents (top), and (i) a snake robot (middle), as well as a hexapod robot~(bottom).
The A3C meta-agent stores a shared policy, that multiple worker agents implement and improve distributedly.
Similarly, a single articulated robot can be separated into independent portions, each controlled by a trained agent learning -- and ultimately implementing -- a common policy in a decentralized manner.
In our example, a snake robot is separated into $6$ control windows, with local shape parameters $A_i,\omega_i,\mu_i$ ($i=1,..,6$).
Similarly, we consider a hexapod robot whose legs are controlled by an agent each, with local parameters $\theta_i,VE_i,\tau_i$.}
\label{TRO2018-SnakeAgent}
\vspace{-0.27cm}
\end{figure}

Our primary contribution shows how a low-level agent in the A3C algorithm can be defined as one of the independently controlled portions of an articulated robot~(Figure~\ref{TRO2018-SnakeAgent}) in a distributed learning framework.

This paper explores the fundamental link between distributed learning approaches and decentralized control architectures, our approach is able to quickly learn near-optimal decentralized policies on a single platform, for an efficient hardware implementation.
Our previous works focused on offline learning of decentralized policies using A3C~\cite{ICRA2018-DistributedLearning}; however, in the present paper, apart from the offline approach, we also extend our approach to online distributed learning of decentralized policies.
We present a case of offline training for a snake robot, and a case of online training directly on hardware for a hexapod robot.
This choice is motivated by our desire to show the feasibility of both training approaches on robots with different morphologies, thus covering the different applications of our approach, without the need to specifically study all the cross cases.
In doing so, we note that online training raises two main challenges: first, since different agents share the same experiences on a single robot, the learning task can be destabilized by the resulting, highly-correlated learning gradients being pushed to the global neural network.
To overcome this problem, we rely on experience replay~\cite{experience} that helps decorrelating the agents' experiences with time during training, similar to our recent multi-robot construction work~\cite{DARS2018-DistributedAssembly}.
Second, an agent's actions may affect the other agents' states during learning, and rewards need to be adapted to truly reflect each agent's contribution to the robot's state.
That is, in this work, we need to explicitly address \textit{credit assignment} to enable consistent~training.

The paper is outlined as follows:
Section~\ref{TRO2018-background} provides a short overview of the A3C algorithm, as well as recent works in multi-agent reinforcement learning (MARL).
In Section~\ref{TRO2018-SEAsnake}, we first summarize the general shape-based controller for serpenoid locomotion, whose feedback controller, which adapts the amplitude and frequency of the snake based on proprioceptive sensory input, we seek to replace by a trained agent.
We then outline the reinforcement problem structure, as well as the agent structure used in this work.
After describing the offline training procedure of our agent, we finally validate the learned policy experimentally on a snake robot and discuss our results.
Section~\ref{TRO2018-snakemonster} follows the same outline as Section~\ref{TRO2018-SEAsnake}, but focuses on learning a decentralized controller online, and directly on hardware, for the stabilized locomotion of a hexapod robot.
The learned controller uses proprioceptive and inertial feedback to adapt the pose of an individual leg for the purpose of leveling the robot's body.
We finally validate our learned policy on a hexapod robot, locomoting on inclined planes as well as in rocks, and show how the learned policy naturally scales to varying robot morphologies and applies to a quadruped without further training.
An overall discussion of our approach is presented in Section~\ref{TRO2018-Discussion}, and concluding remarks and future works in Section~\ref{TRO2018-conclusion}.


\section{Background}
\label{TRO2018-background}


\subsection{Single-Agent Learning}
\label{TRO2018-background-A3C}

Existing learning-based approaches often consider locomotive tasks at the whole robot level~\cite{Peng2016,Yamashina2011,Ito2002}.
These approaches usually train an agent to select actions that will simultaneously affect all the degrees of freedom of the robot, most likely sampled from a large-dimensional action space.
Therefore, these approaches usually require a large amount of experiences to converge to good/near-optimal policies.

Recent advances in the field of deep reinforcement learning (RL) have highlighted the benefits of relying on multiple independent agents to learn a \textit{common}, optimal policy in a distributed -- and thus more time-efficient -- manner.
Some works have focused on learning \textit{homogeneous} decentralized controllers -- i.e., a common policy is enacted in the different portions of the robot based on local feedback -- for locomotion~\cite{ha2018automated,haasdijk2010hyperneat,baldassarre2003evolution}, but have only considered single-agent learning approaches where a single agent learns a controller that is then applied to different portions of the robot; others have focused on learning heterogeneous controllers in a distributed manner~\cite{christensen2013distributed,maes1990learning}.
In either case, however, these approaches did not take advantage of running multiple learning agents to speed up training, even when considering distributed learning.

Mnih et al. proposed a novel single-agent learning approach for deep reinforcement learning~\cite{mnih2016asynchronous}, taking advantage of multi-core architectures to obtain near-linear speed-up via distributed learning, where the learning task is distributed to several agents that asynchronously update a shared neural network, based on their individual experiences in independent learning environments.
That is, these methods rely on multiple agents to distributedly learn a single-agent policy.
A general meta-agent stores a global network, to which the worker agents regularly push their individual gradients during training, and then pull the most recent global weights that aggregate the lessons learned by all agents.

The increase in performance is mainly due to the fact that multiple agents, experiencing different situations in independent environments, produce more decorrelated gradients that are pushed to the global net.
Additionally, agents in different environments can better cover the state-action space, and faster than a single agent (e.g., a deep Q-learning agent~\cite{hausknecht2015deep,mnih2013playing,van2016deep}).
Among these asynchronous methods, the A3C algorithm, extending the standard Advantage Actor-Critic learner to asynchronous distributed learning, was shown to produce the fastest learning rate in a stable manner~\cite{mnih2016asynchronous}.
In this algorithm, the global network has two outputs: a discrete stochastic policy vector (the actor), and a value (the critic), estimating the long-term cumulative reward from the current~state.


\subsection{Multi-Agent Reinforcement Learning (MARL)}
\label{TRO2018-background-MARL}

The first and most important problem encountered when transitioning from the single- to multi-agent case is the curse of dimensionality:
most joint approaches fail as the dimensions of the state-action spaces explode combinatorially, requiring an absurd amount of training data to converge~\cite{Buoniu2010}.
In this context, 
many recent works have focused on decentralized policy learning~\cite{gupta2017cooperative,lowe2017multi,foerster2017counterfactual,foerster2016learning}, where agents each learn their own policy, which should encompass a measure of agent cooperation at least during training.
One way to do so is to train each agent to predict the other agents' actions~\cite{lowe2017multi,foerster2017counterfactual}. 
In most cases, some form of centralized learning is involved, where the sum of experience of all agents is used toward training a common aspect of the problem (e.g., network output or value/advantage calculation)~\cite{gupta2017cooperative,foerster2017counterfactual,foerster2016learning}.
When centrally learning a network output, parameter sharing can be used to let agents share the weights of some of the layers of the neural network and learn faster and in a more stable manner~\cite{gupta2017cooperative}.
In actor-critic approaches, for example, the critic output of the network is often trained centrally with parameter sharing, 
and can be used to train cooperation between agents~\cite{gupta2017cooperative,foerster2017counterfactual}.
Centralized learning can also help when dealing with partially-observable systems, by aggregating all the agents' observations into a single learning process~\cite{gupta2017cooperative,foerster2017counterfactual,foerster2016learning}.
Many of these approaches rely on explicit communication among agents, to share observations or selected actions during training and sometimes policy execution~\cite{lowe2017multi,foerster2017counterfactual,foerster2016learning}.

Second, MARL approaches struggle with stabilizing learning: multiple agents can fail at learning in an ever-changing environment where other agents constantly update their policies
~\cite{gupta2017cooperative,foerster2016learning,littman1994markov,shoham2003multi}.
Recently, actor-critic methods have been experimentally shown to be more robust to changing environments, and to lead to more stable MARL~\cite{gupta2017cooperative,lowe2017multi,foerster2017counterfactual}.
Additionally, for non-Q-learning-based approaches, experience replay can also help stabilize learning~\cite{foerster2017counterfactual,gupta2017cooperative,lowe2017multi}.
Finally, some recent works have proposed curriculum learning, where the complexity of the problem (number of agents in the system) is slowly increased during training, as a way to train numerous agents in a stable manner~\cite{gupta2017cooperative}.


\section{Decentralized shape-based locomotion}
\label{TRO2018-SEAsnake}


\subsection{Background - Shape-based compliant control for articulated locomotion}
\label{TRO2018-SEAsnake-Background}

In this section, we briefly summarize some results from serpenoid locomotion, as well as their use for low-dimensional \textit{shape-based} compliant control of sequential mechanisms~\cite{travers2018shape}.
This is in contrast to non-\textit{shape-based} control strategies, as given in~\cite{ishige2018learning, kano2017tegotae, kano2012local, Kamegawa-2012-107837, liljeback2011experimental}.
Previous works on shape-based control focused on sequential mechanisms as well as legged systems, but we here focus on results pertaining to snake robot locomotion.


\subsubsection{Serpenoid curves}
\label{TRO2018-SEAsnake-Background-Serpenoids}

For an N-jointed snake-like robot, the slithering gait can be parameterized by a sine wave propagating through the lateral plane of the snake (i.e., parallel to the ground)~\cite{Hirose1993,Tesch2009,Whitman2016}. 

\vspace{-0.5cm}
\begin{equation}
\theta (t) = \left( \theta_i (t) \right)_{i=1}^N \mbox{ , with  }
\theta_i (t) \, = \theta_0 + A \sin(\omega_S s_{i} - \omega_{T} t) \\[0.3cm]
\label{TRO2018-SerpenoidCurve}
\end{equation} 
\vspace{-0.7cm}

\noindent where $\theta_i$ are the commanded joint angles, $\theta_0$ the angular offset,
and $A$ the amplitude of the curvature.
$\omega_{T}$ is the temporal frequency of the wave, while $\omega_{S}$ describes its spatial frequency,
which determines the number of sinusoidal periods on the snake robot's body.
$s_{i} \in \{0,2\cdot l_s, \hdots, N\cdot l_s\}$ is the distance from the head to the module $i$, where $l_s$ is the length of one module.




\subsubsection{Shape-based Compliant Control}

Shape-based control makes use of \textit{shape functions} as a method to reduce the dimensionality of high-degree-of-freedom systems~\cite{Travers2015}.
A shape function $h: \Sigma \rightarrow \mathbb{R}^{N}$ determines $\theta(t)$, which is parameterized by a small number of control parameters, namely the shape variables.
Those shape variables define the \textit{shape space} $\Sigma$, which is usually of a lower dimension than $\mathbb{R}^N$.
In our case, we consider the case when the serpenoid curve Eq.\eqref{TRO2018-SerpenoidCurve}'s amplitude and spatial frequency are shape variables.
The shape function $h(A(t),\omega_S(t)) = \theta(t)$ of our system is simply the serpenoid curve Eq.\eqref{TRO2018-SerpenoidCurve}, with the two variables being the amplitude and spatial frequency, while the other parameters are fixed:

\vspace{-0.5cm}
\begin{equation}
\hspace{-0.35cm}\begin{array}{rcl}
h: & \, & \Sigma = \mathbb{R}^{2} \mapsto \mathbb{R}^{N} \\[0.2cm]
h_i(A(t),\omega_S(t)) & = & \theta_0 + A(t) \sin \left( \omega_S(t) \, s_i - \omega_T \, t\right).
\end{array}
\label{TRO2018-ShapeFunction}
\end{equation}

In the state-of-the-art compliant controller for locomotion~\cite{Travers2015}, $A(t)$ and $\omega_S(t)$ (now time-dependent) are determined by the output of an admittance controller~\cite{Ott2010}.
This controller allows the robot to adapt these two serpenoid parameters, with respect to a set of external torques $\mu_{ext} (t) \in \mathbb{R}^{N}$ that are measured at each joint along the snake's body.
This controller enables the robot to comply with its surroundings, by adapting its shape to external forces.
The admittance controller for $\beta = \begin{pmatrix} A(t),\omega_S (t) \end{pmatrix}^T$ reads:

\begin{figure*}[t]
\vspace{0.1cm}
\begin{center}
\includegraphics[width=0.95\linewidth]{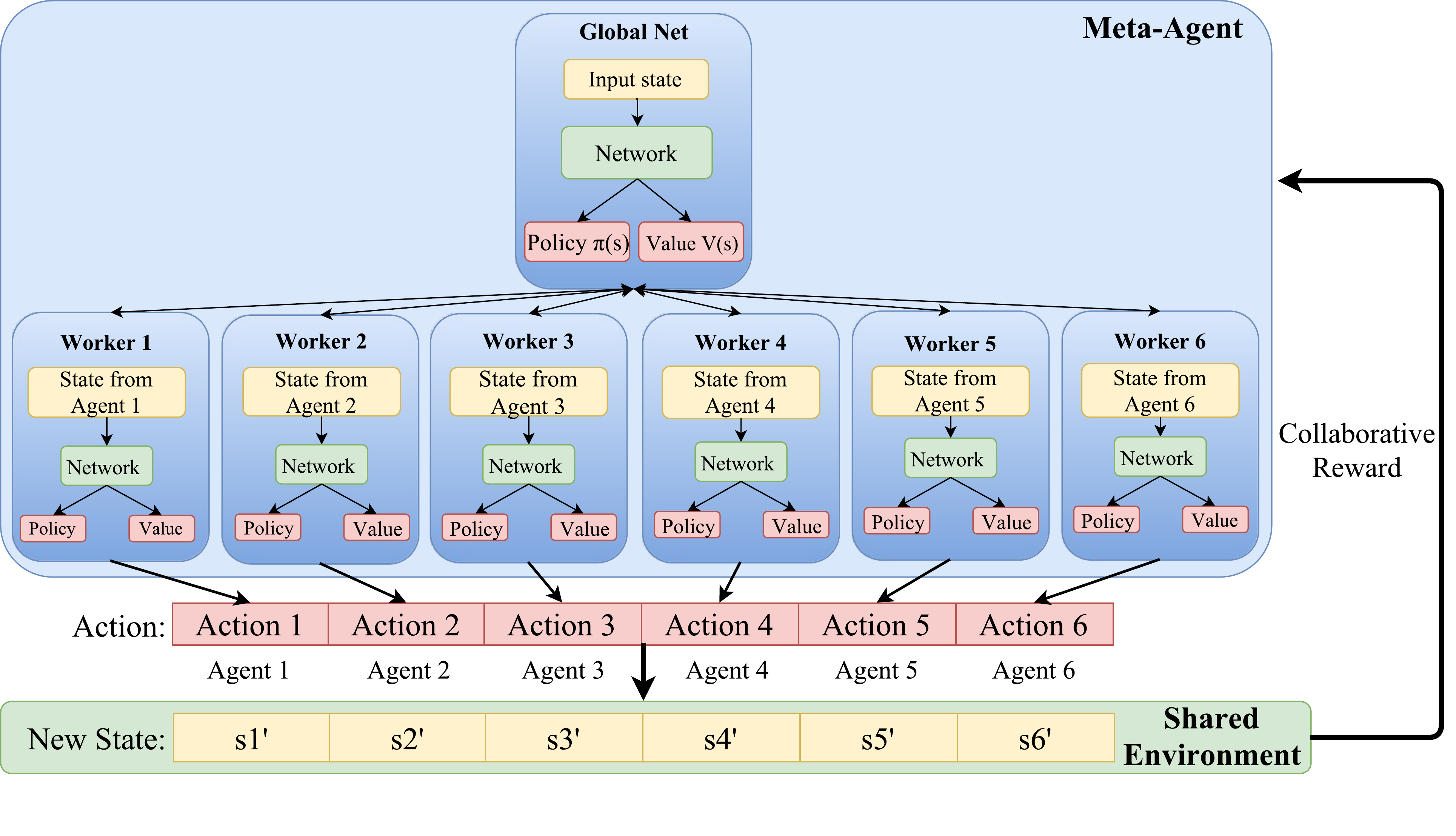}
\end{center}
\vspace{-1.2cm}
\caption{Structure of the A3C Meta-Agent, with global weights for the Actor-Critic network. Each internal agent (worker) updates its local weights based on shared rewards during local interactions with the environment, and regularly pushes its gradients to the global networks.
Each worker draws its state from that of each portion of the articulated robot. Rewards are calculated in response to a vector of the agents' actions that is enacted on the robot.}
\label{TRO2018-AgentGraph}
\end{figure*}

\vspace{-0.3cm}
\begin{equation}
M \, \ddot{\beta}(t) + B \, \dot{\beta}(t) + K \, (\beta(t) - \beta_0) = F (t),
\label{TRO2018-AdmittanceController}
\end{equation}
\vspace{-0.4cm}

\noindent where $M,B,K \in \mathbb{R}^{2 \times 2}$ respectively represent the effective mass, damping and spring constant matrices of the system; $F(t)$ is the external torques $\mu_{ext} (t)$ transformed from the joint space into the shape space (detailed below).

The considered admittance controller Eq.\eqref{TRO2018-AdmittanceController} enables the desired control parameters $\beta$ to respond to an external forcing $F(t)$ with second-order dynamics, acting as a forced spring-mass-damper system.
In the absence of external forces, the control parameters will converge back to their nominal values $\beta_0 = \begin{pmatrix} A_0,\omega_{S,0} \end{pmatrix}^T$.
In this (centralized) controller, where all joints share the same control parameters $A(t)$ and $\omega_S(t)$, the joint angles $\theta_i(t)$ are computed at each time step $t$ from Eq.\eqref{TRO2018-SerpenoidCurve}.

The shape function Eq.\eqref{TRO2018-ShapeFunction} is used to map the external torques measured along the snake's body by the use of the associated Jacobian matrix $J(h) \in \mathbb{R}^{2 \times N}$:

\vspace{-0.18cm}
\begin{equation}
J(h) = \begin{pmatrix}
\frac{\partial h_1(A,\omega_S)}{\partial A} & \cdots & \frac{\partial h_N(A,\omega_S)}{\partial A} \\[0.3cm]
\frac{\partial h_1(A,\omega_S)}{\partial \omega_S} & \cdots & \frac{\partial h_N(A,\omega_S)}{\partial \omega_S}
\end{pmatrix}
\label{TRO2018-Jacobian}
\end{equation}
\vspace{-0.25cm}

External torques are mapped from the joint space to the shape space, via:

\vspace{-0.35cm}
\begin{equation}
F (t) = \left. J(h)\right|_{(A(t),\omega_S(t))} \cdot \mu_{ext} (t),
\label{TRO2018-TorqueMapping}
\end{equation}
\vspace{-0.6cm}

\noindent which is used to update
Eq.\eqref{TRO2018-AdmittanceController}.

In this paper, we seek to replace the admittance controller Eq.\eqref{TRO2018-AdmittanceController}, which iteratively adapts the shape parameters based on proprioceptive sensory feedback, by a trained agent.
In order to simplify the agent's policy and learning process, we leverage the dimensionality reduction offered by the use of shape functions to keep the agent's state space low-dimensional.
That is, we train our agents to reason about and make action in the shape space of the considered robot.


\subsubsection{Decentralized Control}
\label{TRO2018-DecentralizedControl}

Decentralized control extends previous works on shape-based locomotion~\cite{Whitman2016}.
This previous work relies on the use of \textit{activation windows} (i.e., groups of successive joints), and only couples the control of those joints covered by the same window, isolating them from other joints in the snake.
These windows are positioned along the backbone curve of the snake robot in order to create independent groups of neighboring joints, in which torques are sensed and motion parameters are adapted.
This framework allows each separate portion of the snake robot to react independently against local forcing.
Any impact detected by torque sensors locally in a specific section of the body activates a reacting reflex in the individual section only,
resembling biological reflexes where muscle contractions are produced by local feedback~\cite{prochazka1997positive}.

Independent motion parameters for each windows are defined by using sigmoid functions:

\vspace{-0.4cm}
\begin{equation}
\hspace{-0.1cm}\beta(s,t) = \sum_{j=1}^{W}  \beta_{s,j} (t) \bigg[ \frac{1}{1+e^{m(s_{j,s}-s)}} + \frac{1}{1+e^{m(s-s_{j,e})}} \bigg],
\end{equation}
\vspace{-0.3cm}

\noindent where $m$ controls the steepness of the windows, $W$ is the number of windows, and $\beta_{s,j} (t)$ the values of the serpenoid parameters in window $j$ at position $s$ along the snake's backbone.
Window $j$ spans the backbone of the robot over $[s_{j,s}, s_{j,e}] \subset [0;1]$.
In this work, windows are anchored between zero curvature points of the serpenoid curve Eq.~\eqref{TRO2018-SerpenoidCurve}, as is standard practice for use in decentralized shape-based compliant control~\cite{Whitman2016}.
Note that the snake's spatial frequency (determining the number of sine periods along the snake's body) influences the number and positions of amplitude windows along the~backbone~\cite{Whitman2016}.

Recent results have shown that some form of control decentralization is advantageous for a snake robot moving through unknown terrain by relying on local torque-feedback only~\cite{ICRA2018-ProprioceptiveInertial,Whitman2016}, or in combination with normal force-feedback \cite{kano2017tegotae}.
That is, decentralization allows different portions of the snake robot to react independently based on local sensory feedback, in a manner reminiscent of most animals' muscular reflexes.
In particular, these results have considered the case where windows are not fixed to the snake's backbone, but are moved along the snake robot at the same velocity as the serpenoid wave, in order to pass information down the robot's body.
In this paper, we consider the state-of-the-art decentralized control of the snake robot with sliding windows~\cite{Whitman2016}, while shape parameters in each window are updated by a~trained~agent.


\subsection{Policy Representation}
\label{TRO2018-PolicyRepresentation}

In this section, we cast the problem of controlling the snake robot's shape in response to external torques as a Markov decision problem (MDP).
To this end, we detail the state-action space of the learning agents, as well as the structure of the neural nets used to represent the policy agents learn.
We look to the A3C algorithm to distributedly learn a common stochastic policy~\cite{mnih2016asynchronous}, which will then be followed by the joints in each independent window.
In a sense, the A3C meta-agent can represent the whole robot, while the low-level worker agents act as the windows, each selecting local adaptation in the shape parameters and learning from local interactions with the environment, as depicted in Figure~\ref{TRO2018-SnakeAgent}.
Only the global reward, measuring the forward progression of the robot based on the workers' chosen actions, is shared among all workers. 
Alternative reward functions, such as energy efficiency or impact with environment could also result in viable control policies, but were not tested in this paper.


\subsubsection{State Space}
\label{TRO2018-StateSpace}

In reinforcement learning, the state of an agent is the information available to it in order to make its current action decision.
In our case, since this decision is made via the use of a neural network, the state also represents the input to that neural network.
More specifically, the state $\eta$ is a $7$-tuple of shape-based quantities.
It contains the current shape parameters $\beta (s,t)$ of a window on the snake robot (i.e., amplitude and spatial frequency), as well as their nominal values $\beta_0$.
Additionally, the state features the current external torque readings $\mu_{ext} (s,t)$ for the same window, after projection into the shape space by the Jacobian of the system.
Finally, the state needs to feature some information about the cyclic nature of the serpenoid gait.
To this end, we include the \textit{modular time} quantity $[0;1] \ni \tau (t) = \frac{t \, \mod \, T_{s}}{T_{s}}$, which is simply a normalized phase of the serpenoid's wave, with $T_{s} = \frac{2 \pi}{\omega_T}$ the period of the serpenoid curve Eq.\eqref{TRO2018-SerpenoidCurve}.
The state vector $\eta_j (t)$ for window $j \in \{1,..,W\}$ reads:

\vspace{-0.33cm}
\begin{equation}
\eta_j(t) = \langle \tau (t), \, \beta_j (t)^T, \, \mu_{ext} (j,t)^T, \, \beta_0^T \rangle.
\label{TRO2018-StateVector}
\end{equation}
\vspace{-0.48cm}


\subsubsection{Action Space}
\label{TRO2018-ActionSpace}

The admittance controller Eq.\eqref{TRO2018-AdmittanceController} lets the shape parameters evolve with continuous increments.
In this work, we choose to discretize the possible increments in the shape parameters, as a means to reduce the dimensionality of the action space.
Specifically, we let an action be a $2$-tuple $a = \langle a_A, a_\omega \rangle$, with $a_A \in \{-\Delta_A,0,+\Delta_A\}$ and $a_\omega \in \{-\Delta_\omega,0,+\Delta_\omega\}$.
The resulting action space contains 9 discrete actions ($3 \times 3$), which simultaneously update both shape parameters by applying the selected increments:

\vspace{-0.3cm}
\begin{equation}
\beta_j (t+dt) = \beta_j(t) + a(j,t) \cdot dt,
\label{TRO2018-actionApplication}
\end{equation}
\vspace{-0.5cm}

\noindent with $a(j,t) =  \langle a_A, a_\omega \rangle_j$ the action selected by agent $j$ and $dt$ here is set to be $\frac{\pi}{160}$ seconds.


\subsubsection{Actor-Critic Network}
\label{TRO2018-ACNet}

\begin{figure}[t]
\begin{center}
\includegraphics[width=\linewidth]{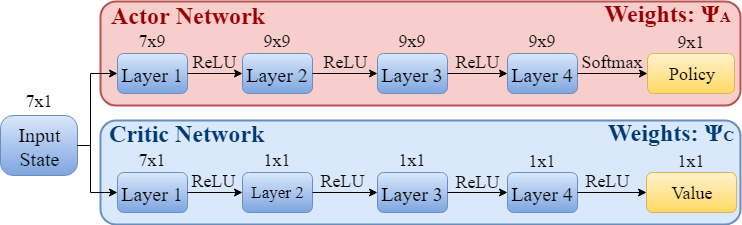}
\end{center}
\vspace{-0.6cm}
\caption{Policy (Actor) and Value (Critic) network used to approximate their respective functions, based on their weights $\Psi_A$ and $\Psi_C$.}
\label{TRO2018-NetworkDiagram}
\end{figure}

We use two deep neural networks with weights $\Psi_A,\Psi_C$ to approximate the stochastic policy function (Actor) and the value function (Critic).
Both networks are fully connected with two hidden layers (see Figure~\ref{TRO2018-NetworkDiagram}).
The output of each layer of the Actor network passes through a rectifier linear unit (ReLU), except for the output layer that estimates the stochastic policy using a Softmax function.
We also use a ReLU at the output of each layer of the Critic network to introduce some non-linearity.
The nonstandard use of a ReLU at the output of the value network is explained by the fact that only positive advantages can be calculated in practice, since the snake robot can only move forward or get stuck (resulting in positive or zero rewards).

Six workers, which correspond to the six independent windows along the backbone of the snake\footnote{As mentioned in Section~\ref{TRO2018-DecentralizedControl}, the current shape of the snake robot influences the number and placement of the independent windows along its backbone.
However, for the implementation of the decentralized controller, we cap the number of windows to $6$. This cap has been selected because the number of windows along the backbone of the snake robot has never experimentally exceeded $5$.
If less than $6$ windows are positioned along the snake's backbone at a given time, the other windows are assumed to be composed of $0$ modules and experience no external torque.
Their shape parameters are additionally assumed to be the nominal ones $\beta_0$.}, are trained concurrently and optimize their individual weights using gradient descent.
Each worker calculates its own successive gradients during each episode.
At the end of each episode, which typically lasts around $89 \, dt$ ($\approx 1.75$ seconds), each worker updates the global network and then collects the new state of the global weights.

We used an entropy-based loss function introduced in~\cite{babaeizadeh2016reinforcement} to update the policy, $\pi(a_t \, | \, s_t;\Psi_A)$:

\newpage $\,$
\vspace{-0.38cm}
\begin{equation}
\begin{array}{lcl}
f_\pi (\Psi_A) & = & \log\left[ \pi(a_t \, | \, s_t;\Psi_A) \right] \left( R_t - V(s_t;\Psi_A)\right) \\[0.2cm]
&  & + \kappa \cdot H\left( \pi(s_t;\Psi_A) \right),
\end{array}
\label{TRO2018-PolicyLoss}
\end{equation}
\vspace{-0.23cm}

In Eq.\eqref{TRO2018-PolicyLoss}, notice that the policy loss is calculated using the \textit{advantage} $\left( R_t - V(s_t;\Psi_A)\right)$, where $R_t$ is the estimated discounted reward over time $t$, instead of only using the cumulative discounted reward $R_t$.
The advantage measures how favorable a given action is in term of long-term expected reward, compared to the value of the current state acting as a baseline of the state's overall quality.
This comparison has been shown to improve a learning agent's ability to not over-/under-estimate the quality of the available actions in very favorable/unfavorable states~\cite{mnih2016asynchronous}, and to improve its decision-making in such difficult states.
$H\left( \pi(s_t;\Psi_A) \right)$ is the entropy factor used to encourage exploration in training and $\kappa \in \mathbb{R}$ helps manage exploration and exploitation, with higher $\kappa$ favoring exploration.
The value loss function reads:

\vspace{-0.28cm}
\begin{equation}
f_v (\Psi_C) = \left( R_t - V(s_t;\Psi_C)\right)^2.
\label{TRO2018-ValueLoss}
\end{equation}
\vspace{-0.44cm}

The stochastic policy is optimized using policy gradient.
During training, we compute the gradients of both functions and optimize the functions using the Adam Optimizer~\cite{Kingma2014}.


\subsubsection{Shared Rewards}
\label{TRO2018-SharedRewards}

Different low-level workers have individual input states, actions and new states corresponding to independent windows.
However, we propose that workers learn based on shared rewards for the combination of their actions.
That is, a shared reward is calculated based on the current progression of the robot, measured from its initial position at the beginning of the current episode $X_{0,i}$ as

\vspace{-0.3cm}
\begin{equation}
r_t = \tanh\left( \lambda_r \cdot \left( \Vert X(t) \Vert_2 - \Vert X_{0,i} \Vert_2 \right) \right),
\label{TRO2018-RewardFunction}
\end{equation}
\vspace{-0.5cm}

\noindent with $X(t)$ the current position of the robot in the world frame and $\lambda_r$ a scaling factor controlling the slope of the hyperbolic tangent.
Note that the reward is normalized by using the $\tanh$ function, to avoid unwanted spikes in rewards, e.g., when the robot suddenly boosts forward due to dynamic effects.

We expect shared rewards to allow the joint actions of all agents to be valued simultaneously, in order to better train them for a collaborative locomotion task. It is also worth noting that we only design and test this shared reward based on the progression as our goal is to optimize the forward progression of the robot. The reward can be designed to optimize the other metrics such as energy and impact.
The global structure of the A3C meta-agent in this paper is illustrated in Figure~\ref{TRO2018-AgentGraph}.



\subsection{Learning}
\label{TRO2018-Learning}

To effectively learn the policy $\pi(a_t \, | \, s_t;\Psi_A)$ using real robot hardware, the state-action space must ideally be densely sampled in the region of realistic parameters.
However, using randomly seeded values to initialize $\pi$ would likely induce the need for thousands of trials to collect enough data to thoroughly sample this space, which is generally not feasible when learning on hardware (as opposed to running thousands of simulated scenarios).
In earlier works, replaying experience from a database to train agents has proven successful in both stabilizing and improving the learned policy~\cite{Bruin2015}.
We use elements of this idea to train our learning agents more effectively.\\
\indent A3C is inherently an on-policy learning algorithm, which can be used to train agents by freely allowing it to sample its state-action space.
If A3C is used off-policy, the gradient updates in the algorithm cannot be theoretically guaranteed to be correct in general.
However, in this work, we experimentally show that A3C can be used to learn off-policy from data sampled by a near-optimal policy.
We believe that this can be the case, since sampling the region of the state-action space close to such a near-optimal policy allows for approximate, yet beneficial updates to the policy.
To this end, we propose to train the agents by replaying past experiences, collected from the state-of-the-art compliant~controller (in a manner resembling imitation learning).\\
\indent In order to build the experience replay database, we performed $310$ trials of the snake robot in randomized peg arrays and with randomized control parameters.
In doing so, we record actions that result in positive and negative reward values for different states (in practice, only positive or zero rewards, as explained in Section~\ref{TRO2018-ACNet}).
In the compliant controller, three $2 \times 2$ diagonal matrices (i.e., a total of 6 parameters) $M$, $B$, and $K$, control the mass, damping, and spring constant of the system.
For each trial, different values for the diagonal terms of $M$, $B$, and $K$ are drawn at random from a uniform distribution over $[\frac12;5]$.
The reward is calculated at each time step using Eq.\eqref{TRO2018-RewardFunction}, by relying on an overhead motion capture system to record the position of the snake robot from tracking beads placed along the snake's backbone.
For each time step, the state and action is recorded
as described in Sections~\ref{TRO2018-StateSpace}-\ref{TRO2018-ActionSpace}.
Each trial is run for approximately 15 seconds in length which we then divided into several episodes. 
After collecting the $310$ trials of the compliant controller in randomized peg arrays (e.g., Figure~\ref{TRO2018-LearnedController}), we stored the data from each trial in a single experience database.

We then assigned a single window to each agent, which was kept constant across all training runs.
During training, each agent asynchronously selected a random run, and then a random starting point in the run.
The agent then learned from an episode of $89$ action-state-reward steps before randomly selecting a new episode.
Leveraging the symmetric nature of the serpenoid curve, we chose $89$ steps to match the length of half of a gait cycle (given our choice of $dt$).
That is, this choice of episode length should provide agents with enough temporally-correlated experiences to connect episodes in a smooth and meaningful manner, and thus learn an efficient global policy.
Each of the agents then sampled episodes until $50,000$ had been drawn.
At this point, we stopped the training process and obtained the learned policy.

\indent The learning parameters used during offline training read:

\newpage $\,$
\vspace{-0.45cm}
\[
\begin{array}{rclrclrcl}
    \Delta_A \hspace{-0.1cm} & \hspace{-0.1cm} = \hspace{-0.1cm} & \hspace{-0.1cm} 0.005 &
    \Delta_\omega \hspace{-0.1cm} & \hspace{-0.1cm} = \hspace{-0.1cm} & \hspace{-0.1cm} 0.012 &\
    \lambda_r \hspace{-0.1cm} & \hspace{-0.1cm} = \hspace{-0.1cm} & \hspace{-0.1cm} 100 \\[0.15cm]
    \gamma \hspace{-0.1cm} & \hspace{-0.1cm} = \hspace{-0.1cm} & \hspace{-0.1cm} 0.995 &
    \quad \alpha_A = \alpha_C \hspace{-0.1cm} & \hspace{-0.1cm} = \hspace{-0.1cm} & \hspace{-0.1cm} 0.0001 &
    \quad \kappa \hspace{-0.1cm} & \hspace{-0.1cm} = \hspace{-0.1cm} & \hspace{-0.1cm} 0.01,
\end{array}
\]
\vspace{-0.3cm}

\noindent with $\gamma$ the discount factor for the policy update, and $\alpha_A,\alpha_C$ the learning rates of the Adam optimization step for the Actor/Critic networks.
The full offline learning code and database used in this work is available online at \url{https://github.com/gsartoretti/Deep-SEA-Snake}.


\subsection{Experimental Validation}
\label{TRO2018-ExperimentalValidation}

\begin{figure}[t]
\begin{center}
\vspace{0.2cm}
\includegraphics[width=\linewidth]{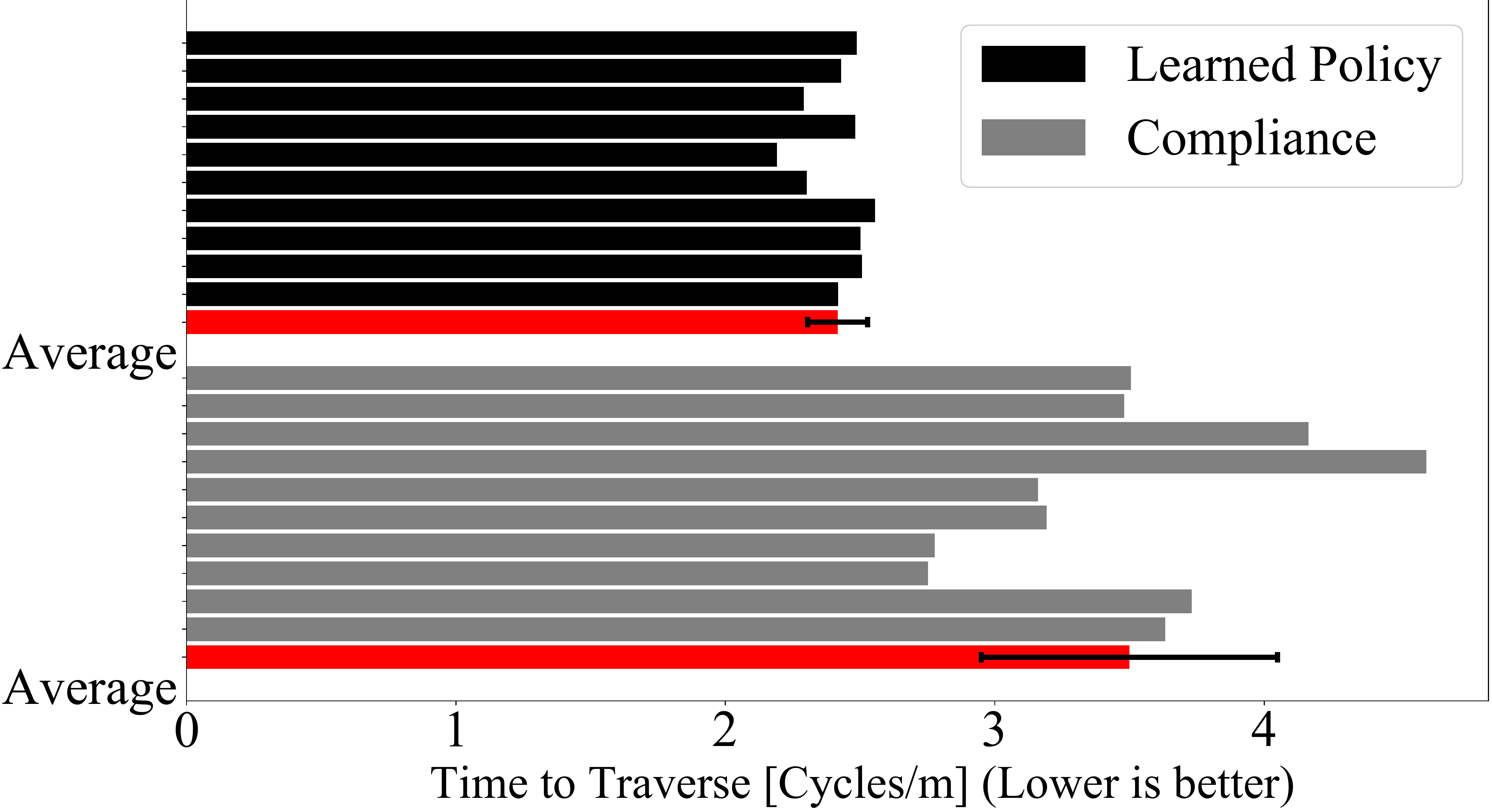}
\end{center}
\vspace{-0.6cm}
\caption{Experimental comparison between both controllers during the $15$ trials, in terms of serpenoid gait cycles needed to progress one meter. Lower values are better. The learning-based controller (black) outperforms the state-of-the-art compliant controller~\cite{Whitman2016} by more than $40\%$.}
\label{TRO2018-ResultsFigure}
\end{figure}

In order to test the validity of our learning approach, we trained a meta-agent to adapt the shape parameters of a snake robot in a decentralized manner.
We implemented the decentralized controller described in Section~\ref{TRO2018-DecentralizedControl} on a snake robot and ran experiments in randomized unstructured environments.
Our goal was to compare the average forward progression of the robot when the shape parameters are updated by the trained agents or the compliant controller Eq.\eqref{TRO2018-AdmittanceController}.


\subsubsection{Experimental Setup}
\label{TRO2018-ExperimentalSetup}

We implemented two versions of the decentralized controller Section~\ref{TRO2018-DecentralizedControl}, using 1) the learned policy, and 2) the compliant controller Eq.\eqref{TRO2018-AdmittanceController} to adapt the shape parameters.
These controllers were tested on a snake robot, composed of sixteen identical series-elastic actuated modules~\cite{Rollinson2014}.
The modules were arranged such that two neighboring modules were torsionally rotated $90$ degrees relative to each other.
The deflection between the input and output of a rubber torsional elastic element is measured using two absolute encoders, allowing us to read the torque measured at each module's rotation axis.
Only the 8 planar modules were active, as only planar gaits were tested.
A braided polyester sleeve covered the snake, reducing the friction with the environment, and forcing the snake to leverage contacts to locomote.
Robot displacement data was collected with an overhead $4$-camera OptiTrack motion capture system (NaturalPoint Inc., 2011).


\subsubsection{Experimental Results}
\label{TRO2018-ExperimentalResults}

\begin{figure}[t]
\vspace{0.18cm}
\begin{center}
\includegraphics[width=0.48\linewidth]{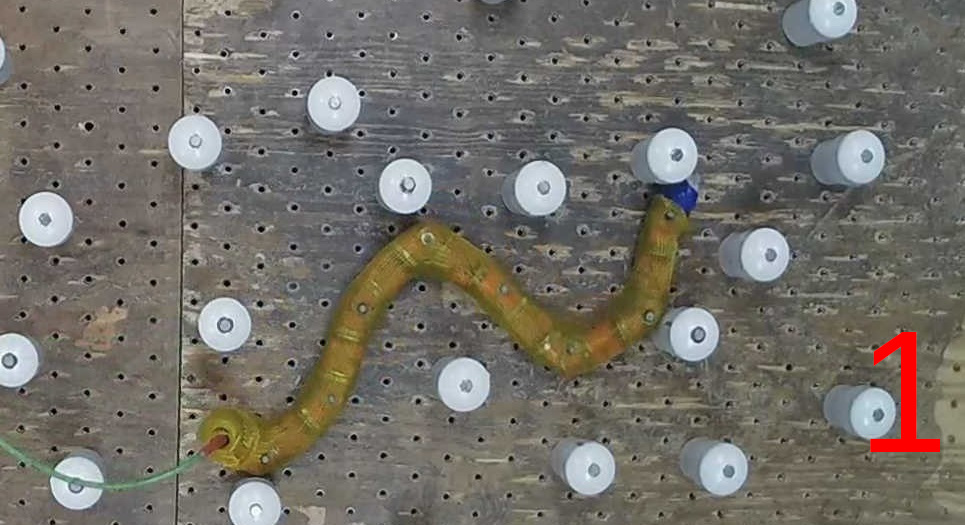} \hfill
\includegraphics[width=0.48\linewidth]{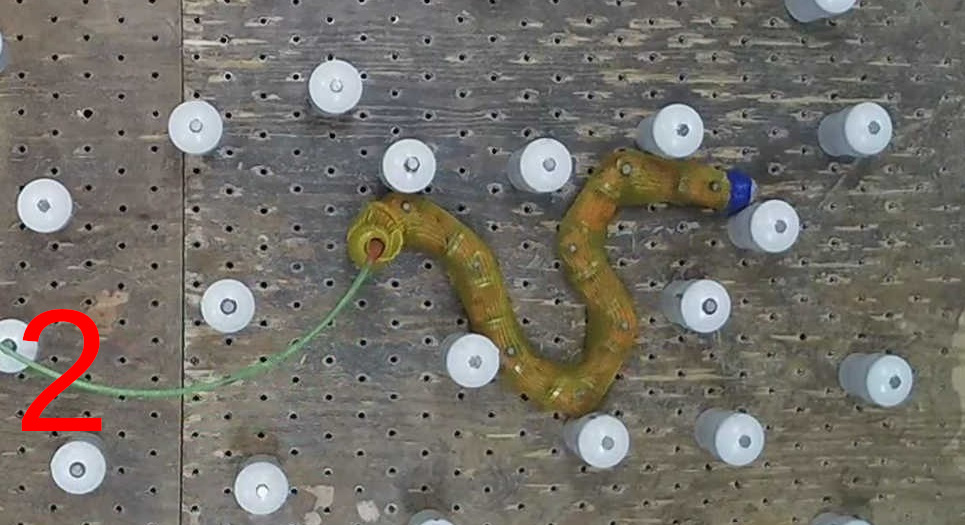} \\[0.19cm]
\includegraphics[width=0.48\linewidth]{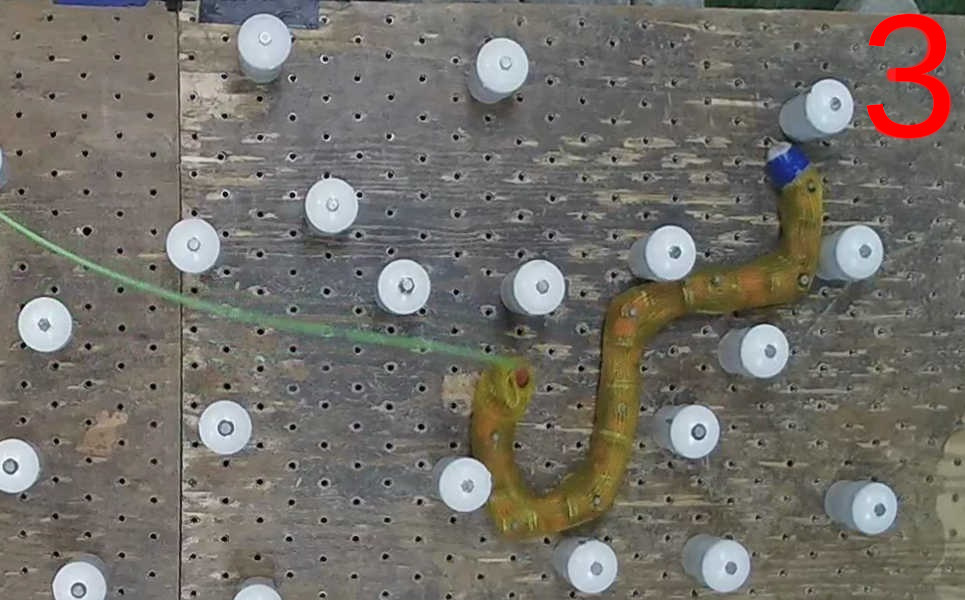} \hfill
\includegraphics[width=0.48\linewidth]{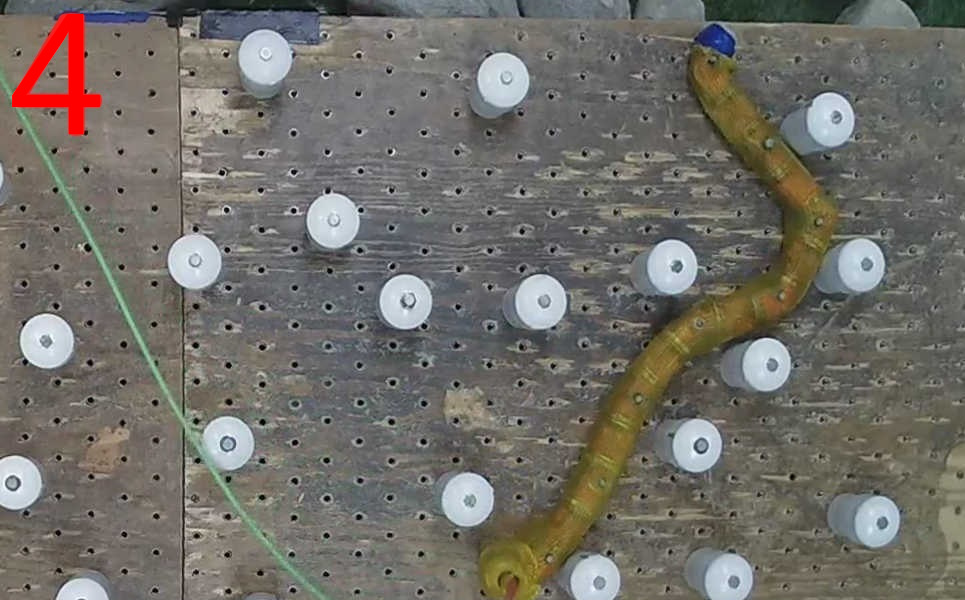}
\end{center}
\vspace{-0.5cm}
\caption{Example frames sampled from a trial, showing the behavior of the learning-based controller. Note how the amplitude and frequency of the waves in each of the windows of the snake robot change to fit the peg array.}
\label{TRO2018-LearnedController}
\vspace{-0.05cm}
\end{figure}

To compare the efficacy of both controllers, we calculate the number of gait cycles needed to traverse one meter and the variance of this number over $15$ trial runs, as established in~\cite{Whitman2016}.
This metric removes the potential variations due to running the gaits with differing temporal frequencies, thus only scoring the controller's kinematic abilities.
Additionally, the variance of the forward progression over the runs serves as an indicator of the repeatability of the controller.
For all the trials, the same unstructured peg array was used for both the compliant controller and the learning-based controller.
Additionally, for each pair of runs, the snake robot was placed in identical starting location as to ensure that both controllers were tested with the same initial configurations.
That is, initial configurations were randomized, but kept identical between both controllers.

Both controllers were run with the same temporal frequency and time step size.
We used the following empirically established optimal parameters for the compliant controller, with time step $dt$.

\vspace{-0.5cm}
\[
\begin{array}{rclrclrcl}
    M \hspace{-0.1cm} & \hspace{-0.1cm} = \hspace{-0.1cm} & \hspace{-0.1cm} \begin{bmatrix}
            1.5 & 0 \\
            0 & 2 \\
        \end{bmatrix} &
    B \hspace{-0.1cm} & \hspace{-0.1cm} = \hspace{-0.1cm} & \hspace{-0.1cm} \begin{bmatrix}
            3 & 0 \\
            0 & 1\\
        \end{bmatrix} &
    K \hspace{-0.1cm} & \hspace{-0.1cm} = \hspace{-0.1cm} & \hspace{-0.1cm} \begin{bmatrix}
            5 & 0 \\
            0 & 1\\
        \end{bmatrix} \\[0.3cm]
    \beta_0 \hspace{-0.1cm} & \hspace{-0.1cm} = \hspace{-0.1cm} & \hspace{-0.1cm} \left[ \frac{\pi}{4}, \, 3\pi \right]^T &
    \hspace{0.2cm} \omega_T^{lat} \hspace{-0.1cm} & \hspace{-0.1cm} = \hspace{-0.1cm} & \hspace{-0.1cm} 1.8 & \quad dt \hspace{-0.1cm} & \hspace{-0.1cm} = \hspace{-0.1cm} & \hspace{-0.1cm} \frac{\pi}{160} [s]
\end{array}
\]
\vspace{-0.28cm}

Over the $15$ trial runs, the learning-based controller achieves an average of $3.27 \, [cycles/m]$ on average, with a variance of $0.65$, compared to the compliant controller, which achieves $4.28 \, [cycles/m]$ on average, with a variance of $1.54$. The results of the compliant controller closely matches, or even slightly outperforms, previous results presented in~\cite{Whitman2016,Travers2015}.
The results are compelling, as they show a distinct increase in the forward progression for the learned policy over the compliant controller.
These results are summarized in Figure~\ref{TRO2018-ResultsFigure}.
The learned policy outperforms the compliant controller by a significant amount (over $40\%$), while the variance is decreased, implying that the learning-based controller is also more repeatable.
Figure~\ref{TRO2018-LearnedController} shows how the learning-based controller modifies the shape parameters in each window during an example trial, in order to adapt to the local configuration of pegs by relying on local force sensing.
Videos of the trials are available online at \url{https://goo.gl/FT6Gwq}.


\section{CPG-based stabilized legged locomotion}
\label{TRO2018-snakemonster}

In the previous section, we considered the offline distributed learning of a decentralized policy, based on experience gathered from a state-of-the-art controller.
Here, we now consider online learning of decentralized policies directly on hardware, by relying on the same distributed learning framework.
To this end, we focus on the problem of stabilizing the body of a hexapod robot during locomotion, by considering each of its legs as one agent, which need to work with the others.
We build upon our recent works on central pattern generator-based locomotion~\cite{ICRA2018-InertialCPG} to let each agent (leg) adapt a single shape parameter controlling the height of a shoulder.
We detail the state-action space and the policy representation used when casting this problem in the RL framework, and propose a reward structure that enables collaborative training.
The trained policy is validated on a series-elastic hexapod robot, while standing and walking on an inclined board and in rocks.


\subsection{Background - CPG Model}
\label{TRO2018-snakemonster-CPGmodel}

We first summarize the central pattern generator (CPG) model, and highlight the model parameters that are adapted to stabilize the robot's body during standing and locomotion.


\subsubsection{Robot Description}
\label{TRO2018-snakemonster-RobotImplementation}

The robot used for our experimental validation is a hexapod made out of series-elastic joints, similar to~\cite{travers2018shape}. These joints are identical to the modules of the snake robot in Section~\ref{TRO2018-SEAsnake}.
Each leg of the robot, shown in Figure~\ref{TRO2018-fig:snakeMonsterACNet} (right), is composed of three modular joints: a planar proximal joint, as well as intermediate and distal vertical joints. The non-joint aluminum portions of the hexapod body (the distal ends of the legs and the rectangular body) are made of hollow aluminum, with an ethernet network switch within the main body.
In this work, we refer to the two proximal joints as the robot's shoulders.


\subsubsection{Mathematical Modelling}
\label{TRO2018-snakemonster-MathematicalModelling}

We express our CPG model as a set of coupled oscillators, i.e., a set of coupled ordinary differential equations, directly in the joint space of the robot.
Let $x(t) = [x_1(t),...,x_n(t)]$ represent the shoulder joint angles of each leg in the axial plane, and $y(t) = [y_1(t),...,y_n(t)]$ those in the sagittal plane.
We consider a general family of parametric limit cycles $H(x,y)$~\cite{SwarmIntelligence2014,ICRA2018-InertialCPG}, taken from the superellipse functions on $\mathbb{R}^2$~\cite{gridgeman1970lame,loria1910spezielle}, and enabling us to vary the shape of the robot's steps during locomotion (see~\cite{ICRA2018-InertialCPG} for details).
The resulting CPG model is:

\vspace{-0.5cm}
\begin{equation}
\hspace{-0.05cm}\left\{\hspace{-0.08cm}
    \begin{array}{lcl}
        \dot{x_{i}}(t) \hspace{-0.12cm} & \hspace{-0.14cm} = \hspace{-0.14cm} & \hspace{-0.14cm} - \omega \cdot \partial H_{y_i} + \gamma \Big(1 - H_{c_i}(x_i(t),y_i(t)) \Big) \partial H_{x_i}\\[0.3cm]
        \dot{y_{i}}(t) \hspace{-0.12cm} & \hspace{-0.14cm} = \hspace{-0.14cm} & \hspace{-0.14cm} + \omega \cdot \partial H_{x_i} + \gamma \Big(1 - H_{c_i}(x_i(t),y_i(t)) \Big) \partial H_{y_i}\\[0.2cm]
        & & \hspace{-0.1cm} + (\lambda\sum_j K_{ij} (y_j(t) - c_{y,j}),
    \end{array}
\right.
\label{TRO2018-eq:CPGsuperOffset}
\end{equation}

\noindent where $\omega$ selects the angular frequency of the gait, $\gamma$ the forcing to the limit cycle, $\lambda$ the coupling strength between legs, and $K$, the coupling matrix~\cite{righetti2008pattern}, defines the gait by controlling the phase relationship between the robots' legs.
The limit superellipse $H(x,y)$, with $\partial H_{\zeta} = \frac{\partial H_{c_i}}{\partial \zeta} (x_i(t), y_i(t))$, reads

\vspace{-0.2cm}
\begin{equation} 
H_c(x,y) = \left | \frac{x - c_{x}}{a} \right |^n + \left | \frac{y - c_{y}}{b} \right |^n.
\label{TRO2018-eq:superElipseOffset}
\end{equation}
\vspace{-0.3cm}

\noindent where $a$ and $b$ select the major/minor axes of the limit ellipse and $n$ defines its curvature level.
For example, $n = 2$ provides a simple elliptic limit cycle; $n > 2$ gives a more rectangular limit cycle with sharper corners as $n$ increases.

Similar to our previous works~\cite{ICRA2018-InertialCPG}, we use the vertical shoulder offsets $c_{y,j} (t)$ to keep the robot's body leveled and at a constant desired height from the ground.
These offsets allow us to modify the pose of the body in the null space of the proximal joint.
That is, we are able to control the body's pitch/roll angles without affecting its yaw (heading).

To implement our CPG model on the robot, we simply output $x_i(t)$ and $y_i(t)$ to the two proximal joints of leg $i$.
A closed-form inverse kinematics function is used to calculate $\theta_{3,i}(t)$, the angle of the distal joint of each leg, during locomotion.
For example, when walking in a straight line, $\theta_{3,i} (t)$ needs to keep the end-effector at a constant, desired distance from the body as the proximal joint angles change.

\begin{figure}[t]
\begin{center}
\includegraphics[width=0.9\linewidth]{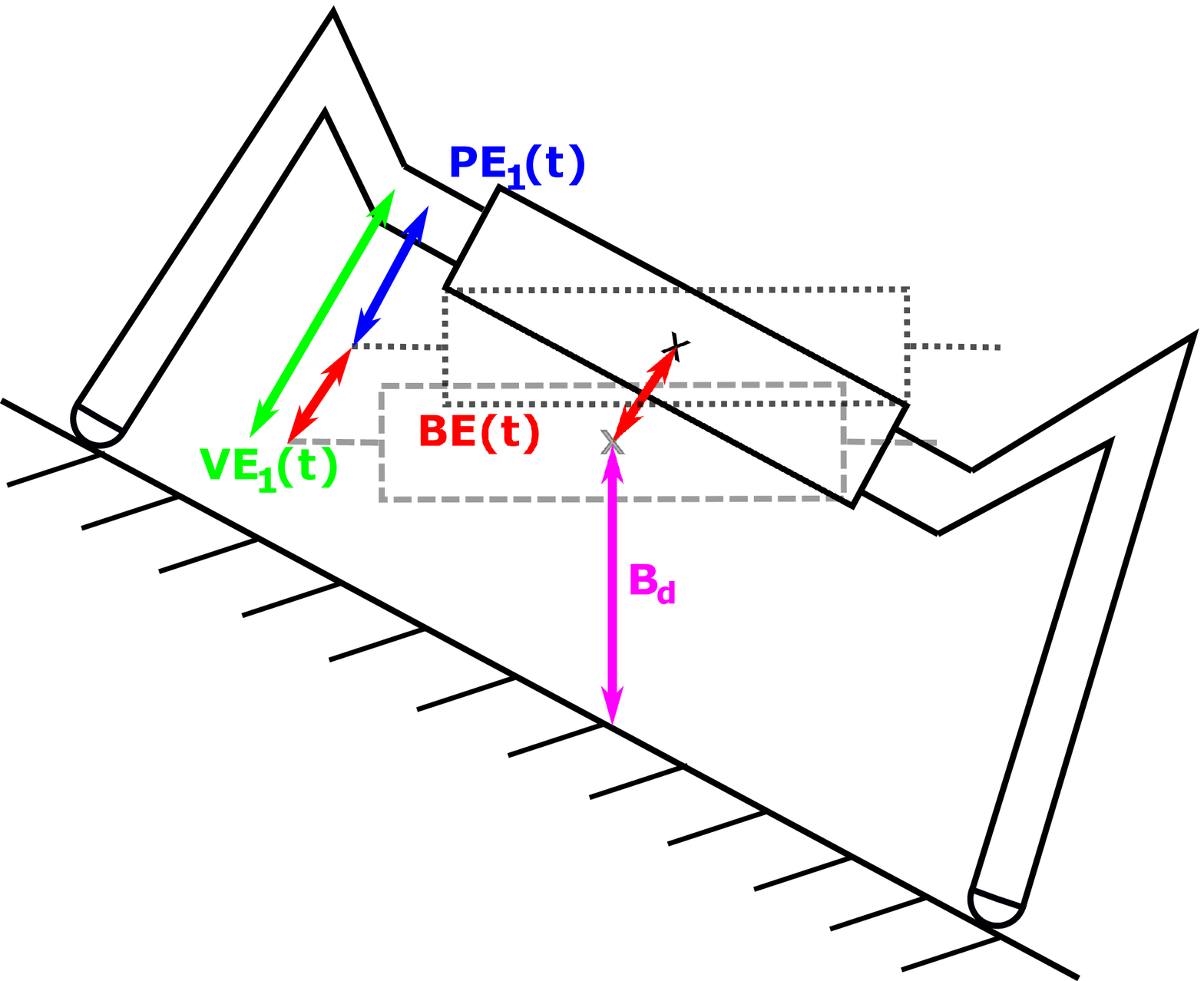}
\end{center}
\vspace{-0.6cm}
\caption{Body posture error $\textcolor{blue}{\textbf{PE}(t)}$, body height error $\textcolor{red}{\textbf{BE}(t)}$, and overall vertical error $\textcolor{green}{\textbf{VE}(t)}$ displayed for agent $1$ (leg 1). The current configuration (i.e., shape) of the robot is displayed in solid lines, corrected body orientation (level orientation of the body) in dotted line and target body pose (level with body height $\textcolor{magenta}{\textbf{B}_d}$ from the ground) in dashed line. The ground plane is estimated from the position of the three grounded feet of the robot (known from the current state of the CPG oscillators).}
\label{TRO2018-verticalError}
\end{figure}


\subsection{Policy Representation}
\label{TRO2018-snakemonster-PolicyRepresentation}

We now describe how stabilizing the hexapod robot is cast as a Markov decision problem (MDP).
To this end, we detail the selected state and action spaces, the policy representation networks, as well as the collaborative reward structure used for distributed~learning.


\subsubsection{State Space}
\label{TRO2018-snakemonster-StateSpace}

The state for each agent (i.e., leg) contains raw information from the sensors of each module of the associated leg, plus a higher-level kinematic information about the overall vertical position error of the same leg.
For each of the three modules of leg $i$ ($1 \leq i \leq 6)$, the state contains the measured joint angle $\theta_{j,i} (t)$ and external torque $\tau_{j,i} (t)$ ($1 \leq j \leq 3)$ at each time step.
The last component of the state vector is the current estimated vertical error $VE_i (t)$ at the proximal joint level.
The vertical error is a sum of the vertical offset induced by any pose error, and any overall body height error. 
Let $SP \in \mathbb{R}^{3 \times 6}$ be a matrix containing the 6 column vectors giving the position of the center of rotation of each of the proximal joints in the robot's body frame.
Given $T \in SO(3)$ the current body pose of the robot, the body posture error $PE (t) \in \mathbb{R}^{1 \times 6}$ reads:

\vspace{-0.2cm}
\begin{equation}
T \cdot SP = \left[ \begin{matrix} XE(t) \\ YE(t) \\ PE(t) \end{matrix} \right].
\label{TRO2018-eq:PoseError}
\end{equation}
\vspace{-0.2cm}

The body height error $BE(t)$ is estimated by first measuring the current height of the robot's body with respect to the ground plane, and then calculating the difference between the current and desired body heights.
To this end, we estimate the ground plane by fitting a plane through the position of the end effector (foot) of the 3 grounded legs (known from the current state of the CPG oscillators).
We note that this ground plane estimation method yields an accurate body height estimation on flat level/inclined ground.
However, the body height may not always be meaningful when the robot locomotes on unstructured environments (such as rock piles), where the body height will be measured from the feet's position and not the actual ground details.
We selected this approach, since there is no general method to estimate the body height of a legged robot locomoting over unstructured terrain, without access to a level map of the terrain, and we show that our approach works very well even on unstructured terrains.
Relevant quantities are illustrated in Figure~\ref{TRO2018-verticalError}.
With this, we express the estimated vertical error $VE (t) \in \mathbb{R}^{1 \times 6}$ as

\vspace{-0.3cm}
\begin{equation}
VE (t) = PE(t) + BE(t).
\label{TRO2018-eq:VerticalError}
\end{equation}
\vspace{-0.5cm}

\noindent The $1 \times 7$ state-vector $s_i(t)$ of agent $i$ at time $t$ finally reads

\vspace{-0.5cm}
\begin{equation}
\langle \theta_{1,i} (t), \theta_{2,i} (t), \theta_{3,i} (t), VE_i (t), \tau_{1,i}(t), \tau_{2,i}(t), \tau_{3,i}(t) \rangle.
\label{TRO2018-eq:stateVector}
\end{equation}

This state space differs from that in Section~\ref{TRO2018-SEAsnake} mainly in that direct joint angles are given to the agent, rather than coordinates in the shape space.
In Section~\ref{TRO2018-SEAsnake}, joint angles could not be used, as there was not a constant number of joints in each window, but the neural net needs a constant number of inputs.
Here, since the number of joints considered by each agent is fixed ($3$ per leg/agent), we directly use those angles in the state space and let agents learn from raw information.

\begin{figure*}[t]
\begin{center}
\includegraphics[height=4.cm]{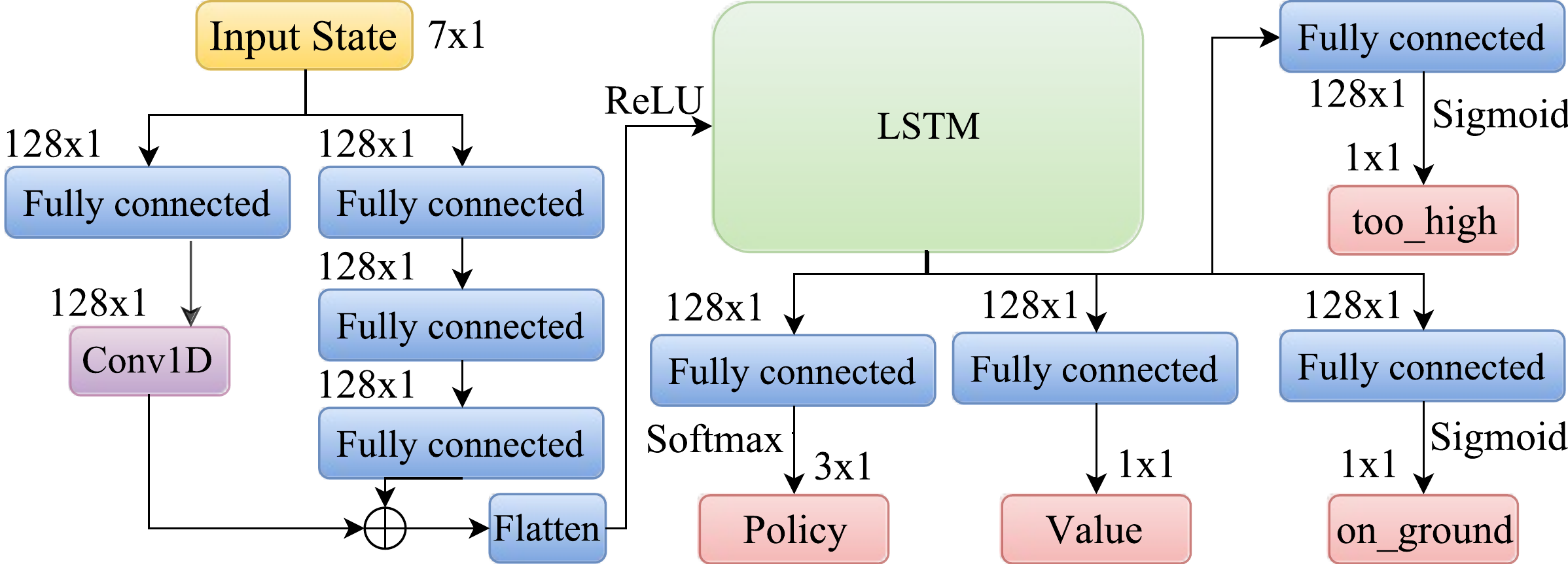} \hfill
\includegraphics[height=4.cm]{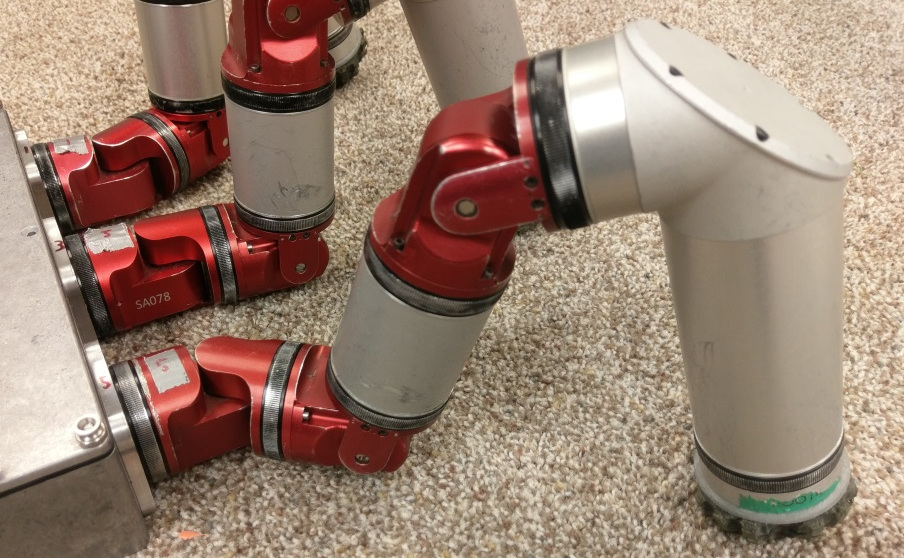}
\end{center}
\vspace{-0.5cm}
\caption{Left: $5$-layer deep neural network used for policy estimation.
Right: Hexapod leg configuration, showing the three joints (red modules).}
\label{TRO2018-fig:snakeMonsterACNet}
\end{figure*}


\subsubsection{Action Space}
\label{TRO2018-snakemonster-ActionSpace}

The action state is composed of a set of discrete positive and negative increments to be applied to current value of the shoulder offsets $c_{y,i} (t)$, given in radians, from Eq.\eqref{TRO2018-eq:CPGsuperOffset}.
Increasing (resp. decreasing) the value of a shoulder offset results in the associated robot's shoulder being moved upward (resp. downward) in the world frame, iff the associated foot is in contact with the ground.

In order to compare between centralized/distributed learning, we need to keep the individual agents' action space low-dimensional, in order to obtain a joint action space as low-dimensional as possible.
For this reason, we propose to only use $3$ actions: $a_i (t) \in \{-1, 0, 1\}$, given in radians, for agent $i$.
For each agent, applying the selected action $a_i (t)$ updates the value of the associated shoulder offset, following:

\vspace{-0.35cm}
\begin{equation}
c_{y,i} (t) = c_{y,i} (t) + a_i (t) \cdot dt,
\label{TRO2018-eq:SMperformAction}
\end{equation}
\vspace{-0.45cm}

\noindent with $dt$ the duration of a time step. This closely mirrors the action space given in the first section, with the difference being that this action is directly in the joint space rather than in the shape space as in the first section.


\subsubsection{Actor-Critic Network}
\label{TRO2018-snakemonster-ACNetwork}

We use a 5-layer fully-connected neural network, each layer with output size 128, followed by a standard long short-term memory (LSTM) layer with 128 outputs, and a residual shortcut~\cite{resnet} between the input layer and the fourth hidden layer (see Figure~\ref{TRO2018-fig:snakeMonsterACNet}).
The output layer consists of the policy neurons with softmax activation, and a value output predicting the reward.
We also added two single output neurons, which enable feature augmentation~\cite{CMU_doom}, to further train the agent on its current state.
More specifically, we let each agent estimate if the robot's body is above the desired body height, as well as whether the associated leg rests on the ground.
Unlike proposed in the ViZDoom task~\cite{CMU_doom}, we connected this layer to the output of the LSTM along with the output layers.
Since the input from the sensors is noisy, training instead the LSTM on the features makes the model more robust against noise.
The primary difference from the network architecture used in Section~\ref{TRO2018-SEAsnake} is the inclusion of an LSTM, which gives the agent the ability to better coordinate its actions across the time domain.
This was not necessary in the network architecture for the snake robot because we made the assumption that the snake robot is purely kinematic, while the hexapod has unmodelled internal dynamics.


\subsubsection{Collaborative Reward Structure}
\label{TRO2018-snakemonster-RewardStructure}

For each agent, we define the reward $r_i (t)$ resulting from enacting the joint actions $a (t) = \langle a_i (t) \rangle_{i=1}^6$, as negative the time-derivative of $PE_i (t)$.
In other words, we favor decreasing the agent-specific shoulder vertical errors with time.
However, simply training agents with this reward structure does not yield a working collaborative policy at the robot level.

During training, some of the legs are sometimes not in contact with the ground.
At these times, the associated agent cannot correct its shoulder height, since increasing/decreasing the shoulder offset of a leg in the air does not change the robot's body pose.
Therefore, the reward of an agent whose associated leg is in the air should not allow any misinterpretation of its actions (since they have no effects).
In this work, we propose to simply set the reward value to $0$ for any agent whose associated leg is currently in the air.
With $\delta_i (t)$ a binary value stating whether leg $i$ contacts the ground, the reward of agent $i$ at time $t$ finally reads:

\vspace{-0.25cm}
\begin{equation}
r_i (t) = r_0 - \delta_i (t) \cdot \frac{ VE_i (t) - VE_i (t-dt) }{dt},
\label{TRO2018-eq:SMreward}
\vspace{-0.1cm}
\end{equation}

\noindent where $r_0$ is negative (in practice, $r_0 = -0.02$), incentivizing agents to stabilize the robot as quickly as possible. This reward structure is fundamentally different from the reward structure used on the snake, as the task is also fundamentally different. However, both reward functions leverage the fact that a majority of the actors performing well correlates with the overall robot performing well.


\subsection{Experimental Validation}
\label{TRO2018-snakemonster-ExperimentalResults}

\begin{figure}[t]
\vspace{-0.1cm}
\begin{center}
\includegraphics[width=\linewidth]{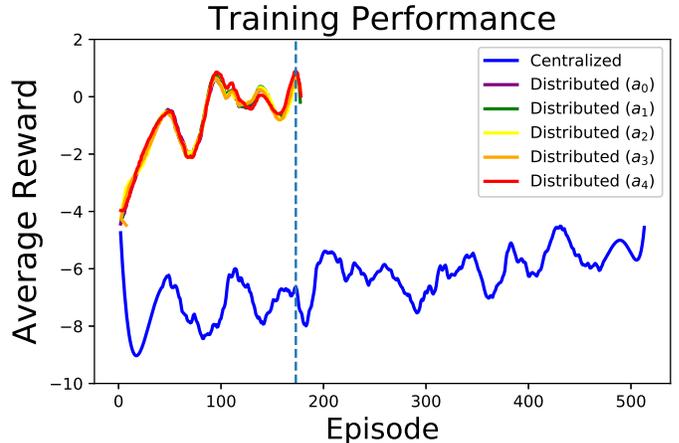}
\end{center}
\vspace{-0.6cm}
\caption{Reward profile during centralized (blue) and distributed (other colors) learning. The centralized approach, despite a network twice as large, did not show any improvement after $500$ episodes due to its $729$ actions.
The distributed approach converges to a smooth policy in $176$ episodes.}
\label{TRO2018-fig:snakemonsterRewards}
\end{figure}

In this section, we describe the experimental learning and validation of our approach on a series-elastic hexapod robot.
We first describe the experimental setting and our learning methodology, and then present and discuss our results.


\subsubsection{Experimental Setting}
\label{TRO2018-snakemonster-ExperimentalSetting}

The experimental setting is composed of a $1.5m \times 1.5m$ flat wooden board covered by a friction-rich carpet, on which the robot stands during training.
The plank is tilted, so as not to let the robot overfit in the presence of a flat environment.
At the beginning of each episode, the robot is initialized with a random pose, with some legs in the air and at least 3 on the ground.
The robot is then given $250$ time steps ($dt = 0.02\,s$) to stabilize; an episode is terminated early if the robot stabilizes within $250$ iterations.
The robot is stabilized if its posture error and its body height error are below a threshold.
Additionally, in order to further vary the challenges given to the agent(s), we rotate the robot by $\frac\pi4$ between successive episodes, to offer varied torque readings to the different legs.
The experimental setting is pictured in Figure~\ref{TRO2018-fig:snakemonsterFrames}. 
To compare the decentralized policy with a centralized policy, we trained a centralized policy in the exact same experimental setting.
As the state space is six times larger, we doubled the neural network's width (number of neurons at each layer) of the decentralized case to still allow for a good function approximation.
The centralized reward structure is such that the average of all legs' individual rewards (Eq.\eqref{TRO2018-eq:SMreward}) is given to the central learning agent.


\subsubsection{Robot Learning}
\label{TRO2018-snakemonster-DistributedLearning}

Since the neural network we proposed is relatively shallow, we trained the agents on a simple desktop computer (6-core AMD Phenom II X6 1055T processor clocked at $2.8$GHz with $8$Gb of RAM).
To decorrelate the gradients from each of the legs, we implement experience replay~\cite{experience} for each agent.
We create an experience pool of the most recent $6$ episodes for each agent.
To train an agent, we randomly draw an experience sample from a randomly selected episode of its pool.
Experiences stored in the pool are different (since we vary the robot's initial poses as explained in Section~\ref{TRO2018-snakemonster-ExperimentalSetting}), thus further reducing over-fitting.
We use the standard Boltzmann action-selection strategy~\cite{sutton1998reinforcement} for the agents during training to encourage exploration.
We stop the training when a smooth policy is found (i.e., the stochastic policy nearly always outputs a single certain action).
The full code used for online learning and offline policy evaluation is available at \url{https://github.com/gsartoretti/TRO2018-hexapodLearning}.


\subsubsection{Results}
\label{TRO2018-snakemonster-Results}

The distributed approach converged to a satisfactory policy after episode 176, while the centralized policy never managed to converge (within 500 episodes) as shown in Figure~\ref{TRO2018-fig:snakemonsterRewards}.
With distributed learning, the performance (reward per episode) immediately started following an increasing trend.
The reward per episode for the distributed policy increased up to episode 176, when it reached $1$ reward per episode, which corresponded to the point at which a smooth policy was found.
With centralized learning, no significant improvement was found even after nearly three times more training episodes.

\begin{figure}[t]
\begin{center}
\includegraphics[width=\linewidth]{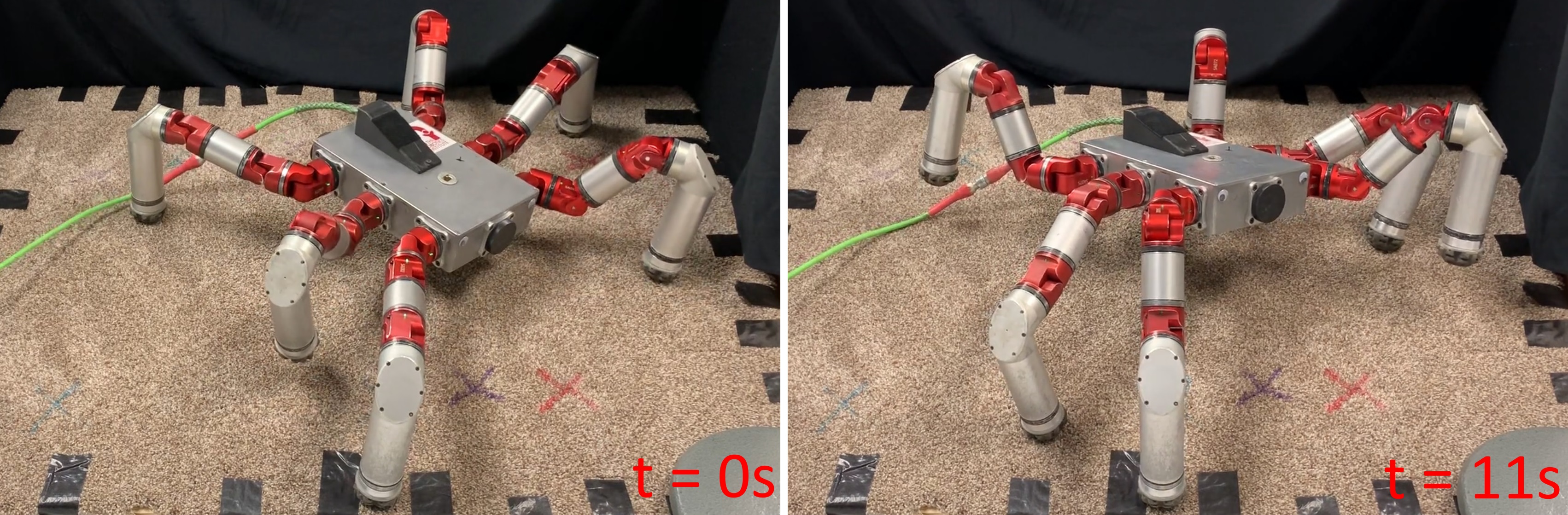} \\[0.04cm]
\includegraphics[width=\linewidth]{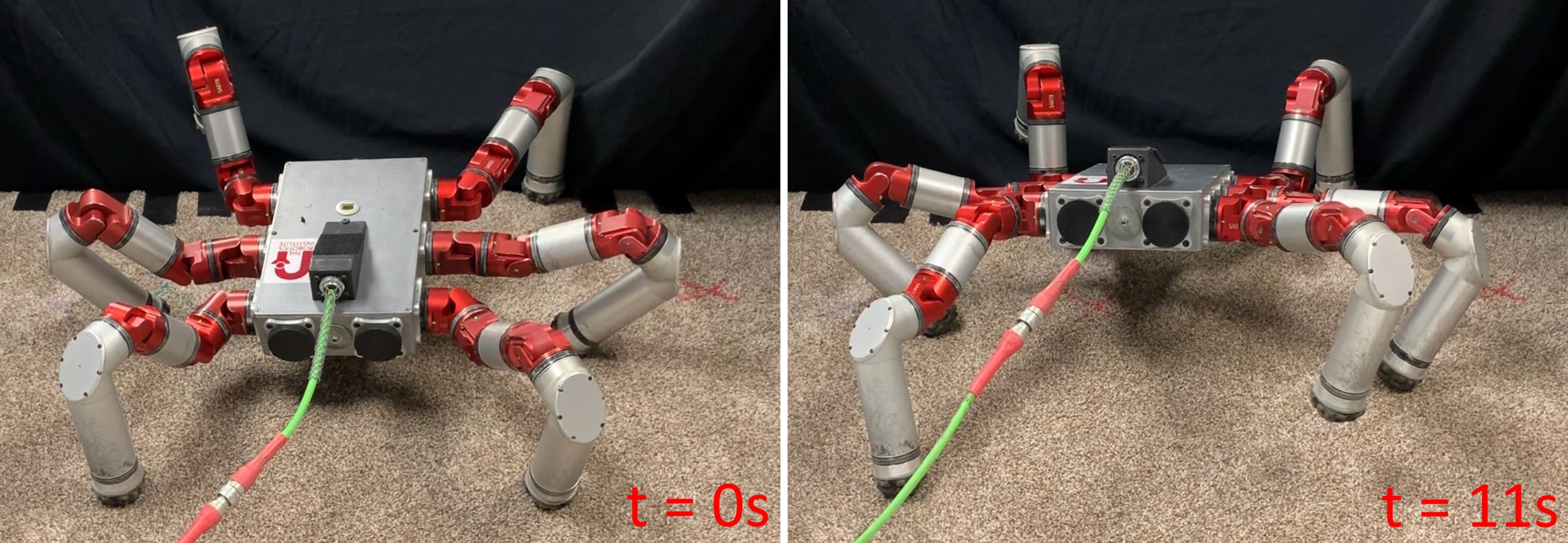}
\end{center}
\vspace{-0.5cm}
\caption{Example start/end robot configuration (up/down rows), showing body stabilization from two random initial poses and respective times to stabilize, using the trained distributed policy.}
\label{TRO2018-fig:snakemonsterFrames}
\vspace{-0.4cm}
\end{figure}

After training, the distributed policy performs as expected, and stabilizes the robot's body.
During execution of the trained policy, we used a slightly different action space $a_i (t) \in \{-0.3, 0, 0.3\}$.
This proved necessary to avoid oscillations around the stable position of the body (i.e., goal body orientation and height), which arose when we allowed the larger increments $a_i (t) = \pm 1$.
We believe that this is due to the fact that an action space composed of two extreme increments and an idle action does not always allow the legs to reach the goal pose of the body, thus creating oscillation patterns around the goal pose.
Note that these oscillations never appeared during training, since episodes were ended when the robot's body reached (close to) the goal pose.
With the action set $\{-0.3, 0, 0.3\}$, stabilization was generally slightly slower, but no oscillations were present when continuously executing the policy.
Figure~\ref{TRO2018-fig:snakemonsterFrames} shows starting and ending frames of two example trials of the distributed learned policy, when the robot is initialized in a random pose on the sloped testing environment and given time to level its body.
All the videos of the training and trials are available at \url{https://goo.gl/Ty7Sta}, showing additional trials in rocky, unstructured terrain.

Once trained, the distributed policy is not limited to a specific number of limbs, and can scale/generalize to a different robot morphology, such as a quadruped or a hexapod where a leg has failed.
Example static stabilization trials on a quadruped robot can be found at \url{https://goo.gl/Ty7Sta}.
The center limbs of each side of the robot body are removed when switching the hexapod to a quadruped. We kept the same CPG parameters but only execute joint angle commands on the four limbs that are attached to the robot body.
This results in static stabilization of the quadruped, but would not allow stable locomotion, for which a new CPG would likely need to be devised (including a new coupling matrix $K$).

\section{Discussion}
\label{TRO2018-Discussion}

In general, we observe that our distributed approach generates control policies which are more scalable and robust to failures (i.e. removing legs from the hexapod robot), and easy to extend by making small modifications to the network architecture.
Additionally, we observe at least equivalent performance with conventional controllers, and in some cases even improved performance.
Due to the nature of deep and convolutional neural networks, our approach should easily extend to incorporating higher dimensional sensors and camera inputs, by simply modifying the network architecture.
By varying the environment in training, a network with enough capacity should also be able to better adapt to unseen, possibly time-varying, more complex environments, but it remains to be seen if training is still viable when accounting for this additional complexity. 
Distributed reinforcement learning approaches seem to offer a way to create control strategies which allow for robust coordination of complex robotic systems without needing to excessively hand-tune the algorithm for a specific robot topology.
Robotic systems which exhibit symmetries should be able to leverage this approach, and therefore benefit from this approach in general, either matching or exceeding performance of classical algorithms while increasing robustness and adaptability.

Our results suggest that our distributed reinforcement learning approach creates efficient snake robot locomotion, meeting or exceeding performance of classical control algorithms such as those given in \cite{Hirose1993, Whitman2016, Tesch2009, Travers2015, kano2012local, travers2018shape, Kamegawa-2012-107837}.
Future work will also compare our learning-based results with other, non-shape-based control approaches, such as~\cite{ishige2018learning,kano2017tegotae}.
In our experiments, we observed that our learned controller reacts to the environment in a similar way to the compliant controller, but visually is able to react more quickly to obstacles, and to hold the frequency and amplitude at set points that allow it to more easily navigate through obstacles.
Additionally, the learned controller seems to better utilize the environment, pushing more strongly against the pegs to create a larger propelling force.

Regarding our hexapedal locomotion results, we first note that the task of leveling the hexapod's body is not only nonlinear, but also very highly dimensional, with multiple paths for achieving stability, greatly complicating the task for multiple agents to reach consensus and perform beneficial actions to stabilize the body.
Since a single agent has no control over the other portions of the robot, and cannot instruct other agents to do any specific action, the difficulties are amplified by the agents' need to collaborate without communication.
Despite these challenges, the proposed framework manages to reach a satisfactory result after relatively few training episodes (corresponding to around $45$ minutes of wall clock time).

In the centralized policy, the challenge of coordination between legs/agents is removed, as a single agent controls the entire robot.
However, due to the inherent difficulty of the task, primarily as a result of the immense action space (729 actions), standard approaches using a centralized controller simply cannot train the agent fast enough.
This is shown in Figure~\ref{TRO2018-fig:snakemonsterRewards}, where even after 500 episodes--corresponding to more than three hours of real world training--the centralized policy had not shown any signs of improvement).
By breaking the state-action space into small regions in which each single agent of the shared environment can learn, we leverage the task homogeneity to distribute the learning, decreasing the total training time and increasing performance.

The learned controller for the hexapod robot is able to closely match the performance of a hand tuned controller given in \cite{ICRA2018-InertialCPG}, quickly flattening the robot body. 
While the performance of the learned controller does not visually seem to exceed that of the hand tuned controller, the learned controller can be more easily modified to account for additional sensory inputs or alternative goals, through the modification of the reward function (such as minimizing the shaking of a mounted camera tightly attached to the robot's body).


\section{Conclusion}
\label{TRO2018-conclusion}

This paper this work explicitly explores the potential relationship between the underlying concepts in distributed learning and decentralized control of articulated robots.
In particular, we consider the problem of adapting the shape parameters of an articulated robot in response to external sensing, in order to optimize its locomotion through unstructured terrain, by casting this problem as a Markov decision problem (MDP).
To solve this MDP, we propose to use a multi-agent extension of the A3C algorithm, a distributed learning method whose structure closely matches the decentralized nature of the underlying locomotion controller.
That is, our approach leverages the decentralized structure of the underlying robot control mechanism, in order to learn a homogeneous policy where a centralized approach would usually not be tractable/trainable.

We first detail how a snake robot can be trained offline using hardware experiments implementing an autonomous decentralized compliant control framework.
There, our experimental results show that the trained policy outperforms the conventional control baseline by more than $40\%$ in terms of steady progression through a series of randomized, highly cluttered evaluation environments (Section~\ref{TRO2018-SEAsnake}).
Note that, for this case of serpenoid locomotion, a comparison with a centralized controller would be more difficult, as the number of control windows on the snake changes with time, making the creation of a centralized controller non-trivial.
Second, we identify learning agents to each of the legs of a hexapod robot, and train agents online to stabilize the robot while standing and locomoting inclined and unstructured terrain.
There, we show how distributed learning enables time-efficient, online learning, whereas a centralized learning approach could not converge due to the associated large-dimensional state-action space (Section~\ref{TRO2018-snakemonster}).
Our results suggest that this approach can be adapted to many different types of articulated robots, by controlling some of their independent parts (e.g., groups of joints, limbs, etc.) in a distributed manner, while fundamentally implementing the same control mechanism in each part.

In general, our distributed approach allows us to learn decentralized policies which naturally exhibit interesting scalability properties.
For example, the policy learned for each leg of the hexapod in Section~\ref{TRO2018-snakemonster} naturally adapts to a quadruped without any additional training.
We envision that this natural scalability could benefit an autonomous legged robot experiencing joint/limb failures in the field, by allowing the other joints/limbs to keep the robot operational.
Based on this observation, we believe that our approach can pave the way to the time-efficient learning of robust, decentralized policies for a wide variety of articulated robots, for use in real-world deployments such as search-and-rescue or inspection. 

Since our current approach generates a homogeneous policy, we believe it can easily generalize to robots whose body segments are not tightly coupled (like the snake robot in this paper) or can be easily decoupled using existing methods (like the hexapod robot in this paper).
However, we acknowledge that the proposed approach cannot be directly applied to some tightly coupled cases where such decoupling strategy may not exist, or where communication of information between limbs/portions of the robot may be necessary.
We further note that, in this work, the way an articulated robot is decoupled into limbs/portions has to be performed manually, based on intuition/experience about the task at hand.
In future work, we will both look at generating potentially heterogeneous policies for different robot segments as well as at learning how to decouple the robot into portions that can be controlled independently.
Additionally, we note that our approach should be able to leverage the use of additional sensory input, and, depending on the task and robot, could improve performance with the addition of cameras, tactile sensors, LiDAR, or other sensors, and we are also interested in exploring these additions in future works.
While explored briefly with the hexapod, we are also interested in maximizing the ability of trained policies to transfer to new, unseen environments, and we plan on exploring how the choice of environment in training and deployment affects the agent.

Finally, in future works, we are also interested in applying this approach to other tasks, such as static or mobile manipulation.
We believe that we can build upon our approach, which relies on centralized learning but decentralized execution, to help address such problems for high-DoF articulated robots.
However, we envision that for such tasks, portions of the robots may need to be more tightly coupled at specific time steps or for specific maneuvers.
In this context, the main challenge will likely be to learn and adapt the level of learning/control decentralization with time, based on the state of the system.


%



\section*{Acknowledgment}

This work was supported by NSF grant 1704256, and by the 2017 CMU RISS and SURF programs.
Detailed comments from anonymous referees contributed to the presentation and quality of this paper.




\bibliographystyle{IEEEtran}
\bibliography{DistributedLearningforArticulatedRobots}

%

\begin{IEEEbiography}[{\includegraphics[width=1in,height=1.25in,clip,keepaspectratio]{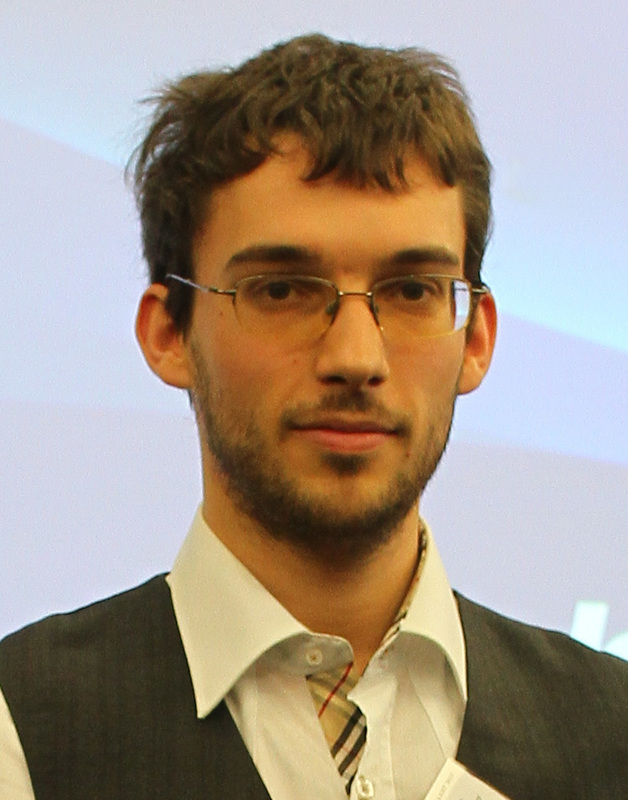}}]{Guillaume Sartoretti}
Guillaume Sartoretti (M'15) is a Postdoctoral Fellow in the Robotics Institute at Carnegie Mellon University, where he works with Prof. Howie Choset. He received his Ph.D. in robotics from EPFL (Switzerland) in 2016 for his dissertation on "Control of Agent Swarms in Random Environments," under the supervision of Prof. Max-Olivier Hongler. He also holds a B.S. and an M.S. degree in Mathematics and Computer Science from the University of Geneva (Switzerland). His research focuses on the distributed/decentralized coordination of numerous agents, at the interface between stochastic modelling, conventional control, and artificial intelligence. Applications range from multi-robot systems, where independent robots need to coordinate their actions to achieve a common goal, to high-DoF articulated robots, where joints need to be carefully coupled during locomotion in rough terrain.
\vspace{-1cm}
\end{IEEEbiography}

\begin{IEEEbiographynophoto}{William Paivine}
William Paivine is an undergraduate student studying computer science at Carnegie Mellon University. He works with Dr. Guillaume Sartoretti and Prof. Howie Choset, with his research interests focusing on the application of distributed reinforcement learning for control of legged robots, artificial intelligence, and planning. 
\vspace{-1cm}
\end{IEEEbiographynophoto}


\begin{IEEEbiography}[{\includegraphics[width=1in,height=1.25in,clip,keepaspectratio]{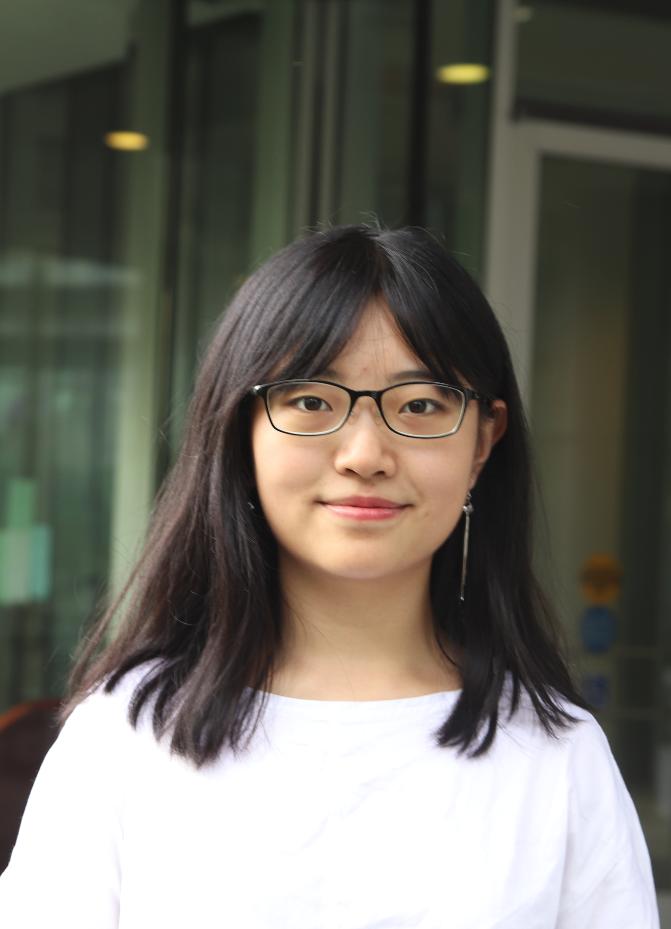}}]{Yunfei Shi}
Yunfei Shi is a Master student in the Robotics Institute at Carnegie Mellon University, where she works with Dr. Guillaume Sartoretti and Prof. Howie Choset. She received her B.Eng. degree in Electrical Engineering from the Hong Kong Polytechnic University. Her research interest lies at the intersection of distributed reinforcement learning, control, and planning.
\vspace{-1cm}
\end{IEEEbiography}

\begin{IEEEbiographynophoto}{Yue Wu}
Yue Wu is a Sophomore Undergraduate student at Carnegie Mellon University, where he worked with Dr. Guillaume Sartoretti and Prof. Howie Choset. He is studying towards a major in Computer Science and a minor in Machine Learning. His research interest lies in on applied distributed reinforcement learning, computer vision, and probabilistic graphical models.
\vspace{-1cm}
\end{IEEEbiographynophoto}


\begin{IEEEbiography}[{\includegraphics[width=1in,height=1.25in,clip,keepaspectratio]{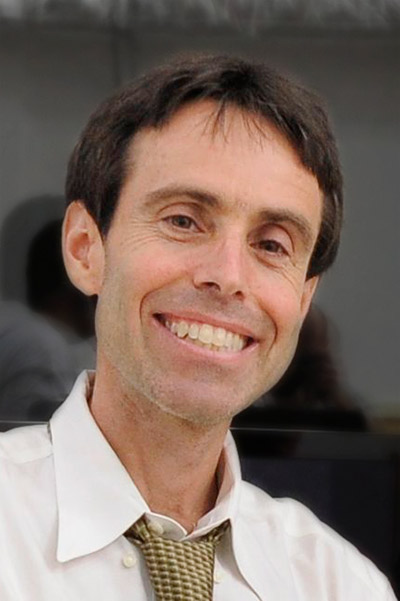}}]{Howie Choset}
(F’15) received the B.S.Eng. degree in computer science, and the B.S.Econ degree in entrepreneurial management from University of Pennsylvania (Wharton), PA, USA, in 1990, the M.S. and Ph.D. degrees in mechanical engineering from California Institute of Technology (Caltech), CA, USA, in 1991 and 1996, respectively. He is a Professor of robotics with Carnegie Mellon University, Pittsburgh, PA, USA. His research group reduces complicated high-dimensional problems found in robotics to low-dimensional simpler ones for design, analysis, and planning. He directs the Undergraduate Robotics Major at Carnegie Mellon and teaches an overview course on robotics, which uses series of custom developed Lego Labs to complement the course work. He has authored a book entitled \textit{Principles of Robot Motion} (MIT Press, 2015).

Prof. Choset was elected as one of the top 100 innovators in the world under 35, by the MIT Technology Review, in 2002. In 2014, Popular Science selected his medical robotics work as the Best of What’s New in Health Care. Finally, his students received the Best Paper awards at the Robotics Industry Association in 1999 and International Conference on Robotics and Automation (ICRA) in 2003. His group’s work has been nominated for Best Paper awards at ICRA in 1997, International Conference on Intelligent Robots and Systems in 2003, 2007, and 2011, and International Conference on Climbing and Walking Robots and Support Technologies for Mobile Machines in 2012; won the Best Paper award at IEEE Bio Rob in 2006, International Symposium on Safety, Security and Rescue Robotics 2012 and 2015; won the Best Video award at International Society for Minimally Invasive Cardiothoracic Surgery 2006 and ICRA 2011; and was nominated for the Best Video award in ICRA 2012.
\vspace{-1cm}
\end{IEEEbiography}




\end{document}